\documentclass[10pt,twocolumn,letterpaper]{article}

\usepackage{cvpr}              
\usepackage{cuted}
\usepackage{capt-of}
\usepackage{multirow}
\usepackage{colortbl}
\usepackage{makecell}
\usepackage[table]{xcolor}
\definecolor{myblue}{RGB}{210, 210, 255}

\definecolor{cvprblue}{rgb}{0.21,0.49,0.74}
\usepackage[pagebackref,breaklinks,colorlinks,allcolors=cvprblue]{hyperref}

\def\confName{CVPR}
\def\confYear{2027}

\title{GenOne: Generate Once, Render Multiple Views for Multi-Camera Driving Video Generation}

\newcommand\blfootnote[1]{%
  \begingroup
  \renewcommand\thefootnote{}\footnote{#1}%
  \addtocounter{footnote}{-1}%
  \endgroup
}

\author{
\fontsize{11pt}{13pt}\selectfont Tianyu Liu$^{1,2}$, Kunming Luo$^1$, Weitao Xiong$^3$, Yingshuang Zou$^1$, Jiahao Lu$^1$, Jiayi Xu$^{1,2}$, Zekai Gu$^1$,\\[2pt]
\fontsize{11pt}{13pt}\selectfont Haiping Wang$^1$, Wei Wu$^2$, Maosheng Ye$^2$, Tongyi Cao$^2$, Yuan Liu$^{1,\dagger}$, Ping Tan$^{1,\dagger}$\\[5pt]
\normalsize $^1$The Hong Kong University of Science and Technology \quad $^2$DeepRoute.AI \quad $^3$Tsinghua University
}

\begin{document}
\maketitle

\setlength{\stripsep}{2pt}
\begin{strip}
    \vspace{-35pt}
    \centering
    \includegraphics[width=1.0\linewidth]{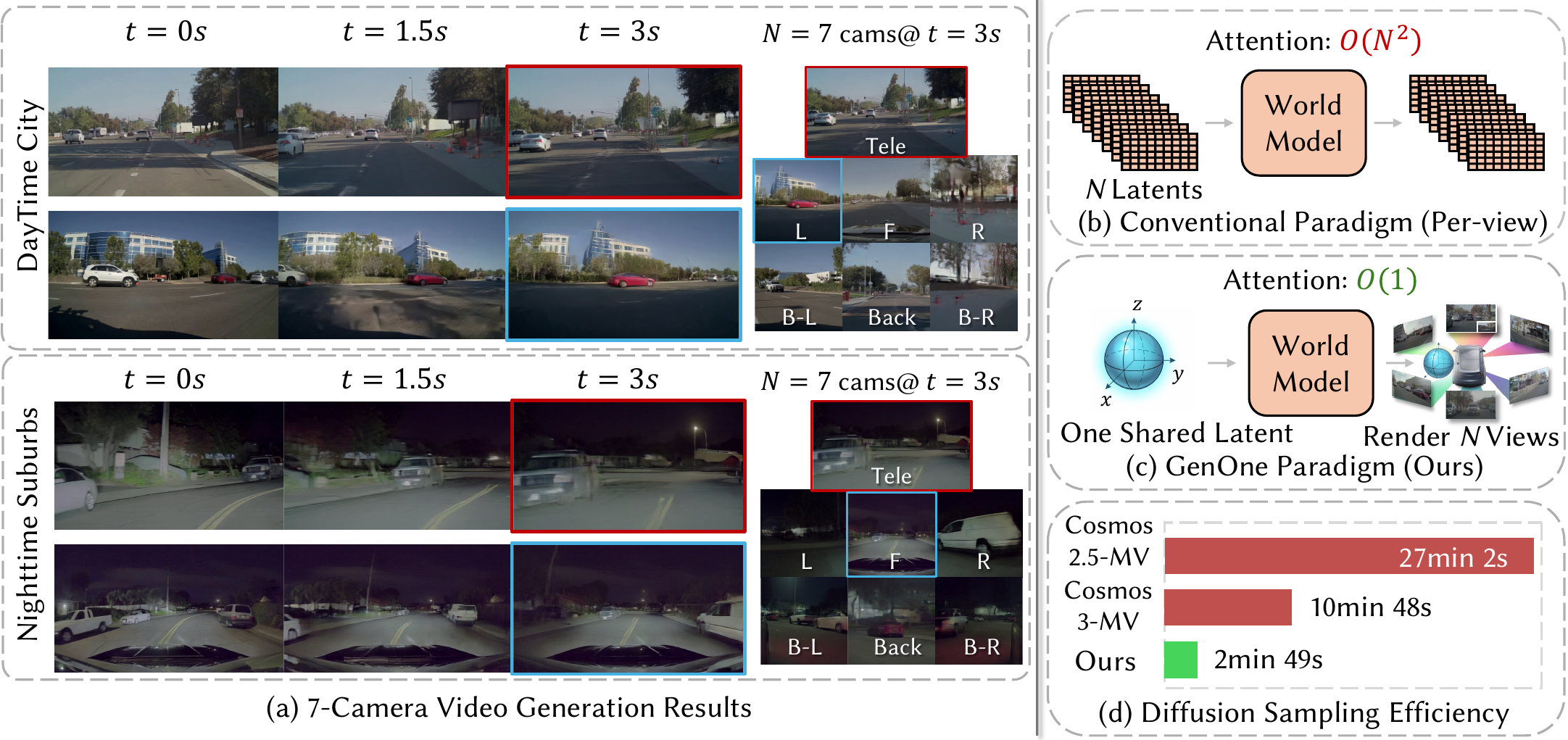}
    \captionof{figure}{\textbf{Generate once, render multiple views.} (a) Daytime and nighttime multi-camera video generation examples conditioned on the initial camera frames. Two camera views are shown at selected video times $t=0$, $1.5$, and $3$s. The grid shows all $N=7$ camera views at \(t=3\)s; red and blue outlines link the two temporal rows to their corresponding views in the grid. Here \(t\) denotes video time and \(N\) the number of output cameras. (b) Dense self-attention over $N$ per-view latent sequences has $O(N^2)$ cost with respect to the camera count. (c) GenOne denoises one shared scene representation and then renders the requested views; for a fixed video duration, its denoiser self-attention cost is $O(1)$ with respect to $N$. (d) Diffusion sampling GPU time, summed across the GPUs used by each method (Tab.~\ref{tab:inference_performance}).}
    \label{fig:teaser}
\end{strip}

\blfootnote{$^\dagger$Corresponding authors. }
\blfootnote{Project Page: \url{https://lty2226262.github.io/genone/}}

\begin{abstract}

Multi-camera video generation is essential for driving world models. However, existing methods typically generate a separate video representation for each camera, repeatedly synthesizing overlapping scene content while requiring additional mechanisms to maintain cross-view consistency, which introduces substantial redundancy and computational overhead. We present GenOne, a \textit{generate once, render multiple views} framework that replaces separate camera-specific video representations with a compact, unified global scene-token representation. By representing shared scene content only once, GenOne greatly reduces the number of tokens required for generation, decoupling the generative cost from the number of output cameras.
To achieve this, we develop a scene tokenizer that consolidates multi-camera videos into shared scene tokens, which can be decoded into multiview videos. We further train a flow-matching model to generate these global scene tokens for efficient multiview video generation. Experiments demonstrate that GenOne substantially accelerates multi-view video generation with competitive visual quality.
\end{abstract}
    
\section{Introduction}
\label{sec:intro}

As autonomous driving shifts toward end-to-end systems and world models~\cite{hwang2024emma,wang2025drivegen,yang2023unisim,agarwal2026cosmos}, high-quality multi-camera video generation has become essential for simulating realistic training and evaluation environments. This demand has driven rapid progress in both multi-camera controllable scene synthesis~\cite{li2024drivingdiffusion,wen2024panaceaplus,zhao2025drivedreamer,gao2025magicdrive,wu2025drivescape,ren2025cosmosdrivedreams,zhang2026anyscene} and multi-camera world models~\cite{wang2024drivewm,lu2024wovogen,russell2025gaia,cheng2026m4world,zhu2026xiaomiworld2,zheng2026xworld16}.

Despite these advances, most existing methods maintain a separate video representation for each camera and exchange information across different cameras to maintain multiview consistency~\cite{gao2025magicdrive,wu2025drivescape,russell2025gaia} (Fig.~\ref{fig:teaser}(b)). Consequently, this leads to a rapid growth of generated tokens for more cameras, even when all these views contain the exact same roads and vehicles. This redundancy is compounded at every diffusion step, making multi-camera video generation costly. In our evaluation, the 35-step Cosmos 2.5-MV and Cosmos 3-MV baselines each use eight NVIDIA H20 GPUs and require 27 min 2 s and 10 min 48 s of aggregate diffusion-sampling GPU time, respectively, to generate a three-second multi-camera video.

Our key insight is that all cameras observe the same driving scene from different viewpoints. This motivates our \emph{generate once, render multiple views} paradigm (Fig.~\ref{fig:teaser}(c)): the world model generates a shared scene representation instead of separate video representation for each camera. These shared tokens model a single world evolving over time; a decoder then renders it into videos at the requested camera viewpoints. For a fixed clip duration, additional cameras therefore increase rendering work but leave the number of tokens processed by the generator unchanged.

To implement this paradigm, we introduce \textbf{GenOne}. We first learn the shared scene representation through multi-camera video reconstruction. Its Q-Former encoder~\cite{li2023blip} aggregates synchronized camera observations into a temporally ordered sequence of shared scene tokens. On top of these scene tokens, our decoder combines SRT-style ray conditioning~\cite{sajjadi2022scene} with timestamp conditioning to predict video latents, resulting in multiview-consistent videos on different cameras. Beyond computational efficiency, this formulation enables flexible multi-camera video synthesis: the same scene-token sequence can be rendered into different numbers of camera streams at different resolutions. Reconstructing all streams from one sequence encourages the scene tokens to capture shared scene content and its changes over time, potentially improving multiview consistency.

Implementing this efficient shared scene representation is non-trivial. The introduced information bottleneck of compact scene tokens limits the capacity to model all scene elements with equal fidelity in the resulting videos. Moving vehicles are particularly important in driving videos because their positions and visibility change across frames and camera views~\cite{yang2023unisim,wen2024panaceaplus}. We therefore introduce a region-aware capacity allocation mechanism in GenOne to prioritize vehicle information within the fixed scene-token budget. Specifically, we project annotated 3D object boxes into each camera view and construct a soft Gaussian obstacle mask from their projected locations and extents. We concatenate each mask with the corresponding video latents before scene encoding, giving the encoder an explicit cue to foreground regions. The same masks also weight the RGB reconstruction loss, so errors in those regions receive greater penalty. By guiding both encoding and reconstruction, this mechanism encourages the shared scene tokens to retain vehicle appearance as the scene evolves over time.

With the shared representation in place, we train a conditional flow-matching model to sample new scene-token sequences, each representing a possible evolution of the driving scene. 
Experiments show that GenOne supports faithful multi-view reconstruction and cross-view consistency while generating videos of competitive visual quality. The architectural gain is clearest at sampling time: under the same 35-step setting, sampling takes 2 min 49 s rather than 29 min 14 s for the per-view basline, a $10.4\times$ speedup. By decoupling scene generation from view rendering, GenOne breaks the link between output camera count and the cost of sampling when generating a new scene.

\section{Related Work}
\label{sec:related_work}

\textbf{Scene Representations and Ray-Conditioned Rendering.} A line of work separates scene encoding from target-view rendering. SRT~\cite{sajjadi2022scene} queries latent scenes with camera rays; the encoder--decoder variant of LVSM~\cite{jin2025lvsm} aggregates views into scene tokens; and RayZer~\cite{jiang2025rayzer} estimates camera parameters from unposed images for ray-conditioned rendering. LagerNVS~\cite{szymanowicz2026lagernvs} uses geometry-informed features for novel-view synthesis, while GLD~\cite{jang2026repurposing} performs multi-view diffusion in geometric foundation-model features. For dynamic scenes, 4DGT~\cite{xu20254dgt} and MoVieS~\cite{lin2026movies} model time-varying geometry and appearance with Gaussians, and TrackTok~\cite{pham2026tracktok} learns persistent object-centric tokens. Closer to our setting, SceneTok~\cite{asim2026scenetok} learns scene tokens for novel-view rendering and generation, while 3DRAE and 3DDiT~\cite{wei20263drae} generate 3D-grounded tokens with flow matching. These methods support changing viewpoints from shared latents but do not explicitly model scene content evolving over time. GenOne instead learns and generates a shared representation of synchronized multi-camera video, queried by camera rays and timestamps.

\begin{figure*}
    \centering
    \includegraphics[width=0.9\linewidth]{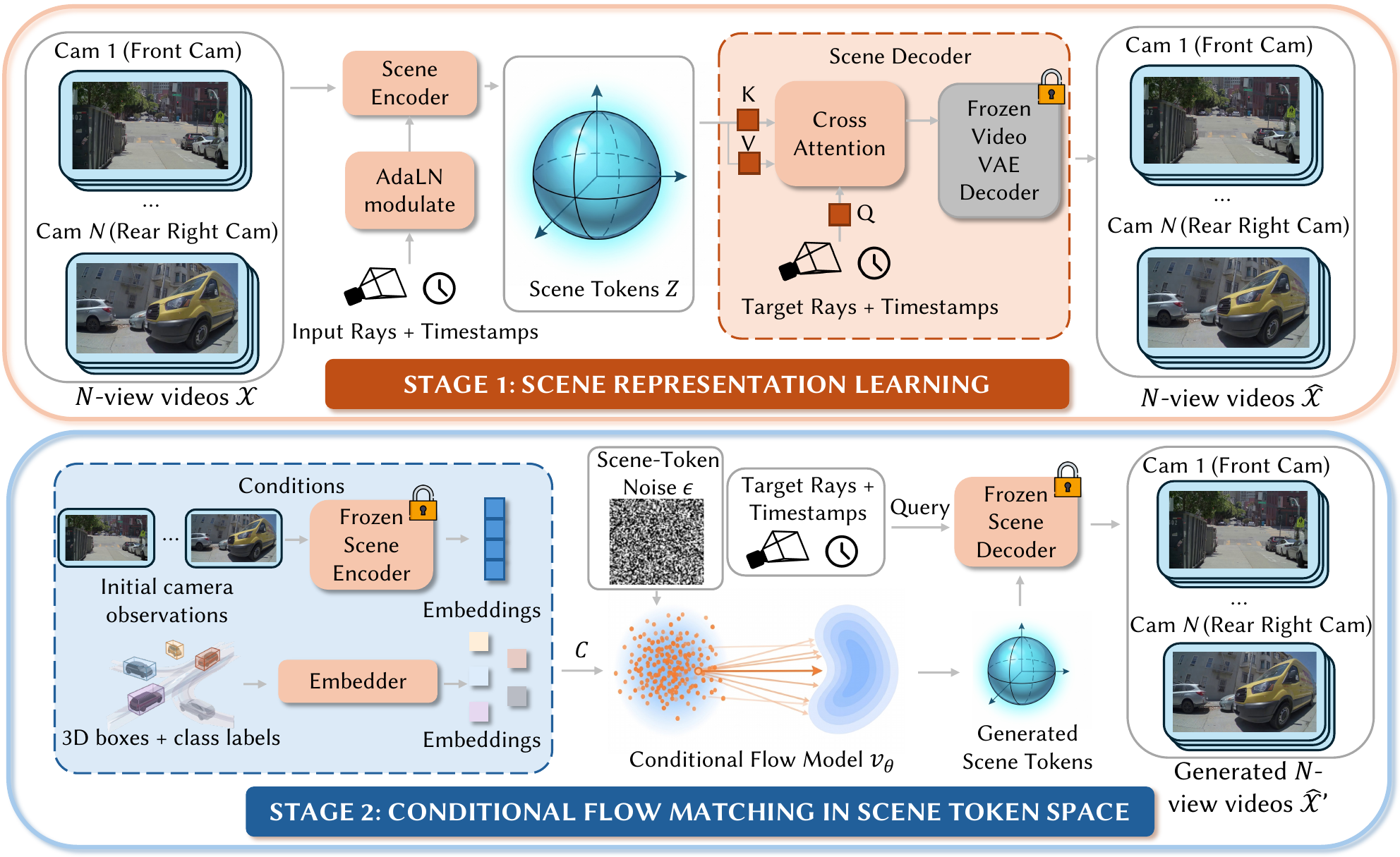}
    \caption{\textbf{Overview of GenOne.} The top and bottom rows show Stage 1 and Stage 2, respectively. In Stage 1, the scene encoder aggregates information from the $N$ input camera videos $\mathcal X$, modulated by input-camera ray and timestamp embeddings via adaptive layer normalization (AdaLN), into a shared scene representation $Z$. For each target view, the scene decoder uses target-ray and timestamp embeddings as queries and attends to $Z$ as keys and values to predict video latents, which the frozen video VAE decoder converts into reconstructed videos $\hat{\mathcal X}$. In Stage 2, the condition $C$ combines tokens from initial camera observations encoded by the frozen scene encoder with embeddings of 3D boxes and class labels. The conditional flow model maps Gaussian noise $\epsilon$ and $C$ to generated scene tokens, which the frozen scene decoder renders into the requested views using their rays and timestamps. Padlock icons indicate frozen modules.}
    \label{fig:main}
\end{figure*}

\noindent\textbf{Multi-Camera Driving Videos Generation.}
MagicDrive-V2~\cite{gao2025magicdrive}, \mbox{GAIA2}~\cite{russell2025gaia}, \mbox{X-World}~\cite{zheng2026xworld16}, and multi-view Cosmos~\cite{ali2025world,agarwal2026cosmos} maintain per-camera video-latent sequences, so their generative token counts remain tied to the requested views. Shared driving representations take different forms: BEVWorld~\cite{zhang2024bevworld} predicts BEV latents for image and LiDAR decoding; \mbox{BEV-VAE}~\cite{chen2025bev} uses a unified BEV latent for layout-controlled multi-view image reconstruction and generation; DriveTok~\cite{zhuo2026drivetok} learns scene tokens with image, depth, semantic, and occupancy supervision; and DriveUniTok~\cite{xu2026driveunitok} combines appearance, projected depth, and ego motion for prediction and planning. AnyScene~\cite{zhang2026anyscene} generates semantic occupancy before camera videos. GenOne instead learns temporally ordered scene tokens through multi-camera video reconstruction and generates them without a BEV grid or intermediate occupancy stage.

\noindent\textbf{Efficiency and Temporal Rollouts.} DiVE~\cite{jiang2026dive} accelerates multi-view sampling through guidance distillation and progressive resolution. FAR-Drive~\cite{li2026far} uses frame-level autoregression for closed-loop interaction; HorizonDrive~\cite{zhang2026horizondrive} addresses rollout drift and distills an efficient student; and OmniDreams~\cite{basant2026omnidreams} targets real-time, action-conditioned simulation. GenOne reduces the representation cost at each flow step by processing one shared scene-token sequence rather than separate sequences for the requested cameras.
\section{Method}
\label{sec:method}

\subsection{Overview}

GenOne has two stages: learning a scene representation (Sec.~\ref{sec:representation}) and generating videos in that representation (Sec.~\ref{sec:generation}). Figure~\ref{fig:main} summarizes GenOne's two-stage pipeline. In the first stage, we train an encoder--decoder to learn a shared scene-token latent space through multi-camera video reconstruction. In the second stage, we freeze the encoder--decoder and train a conditional flow-matching model to generate scene-token sequences in this space.

\subsection{Learning a Scene Representation}
\label{sec:representation}

Learning a scene representation for GenOne raises two questions: (1) how to keep the scene-token count independent of the number of camera views; and (2) how to render requested views from those tokens using camera rays and timestamps. We address these through the encoder and decoder respectively.

\begin{figure}
    \centering
    \includegraphics[width=1.0\linewidth]{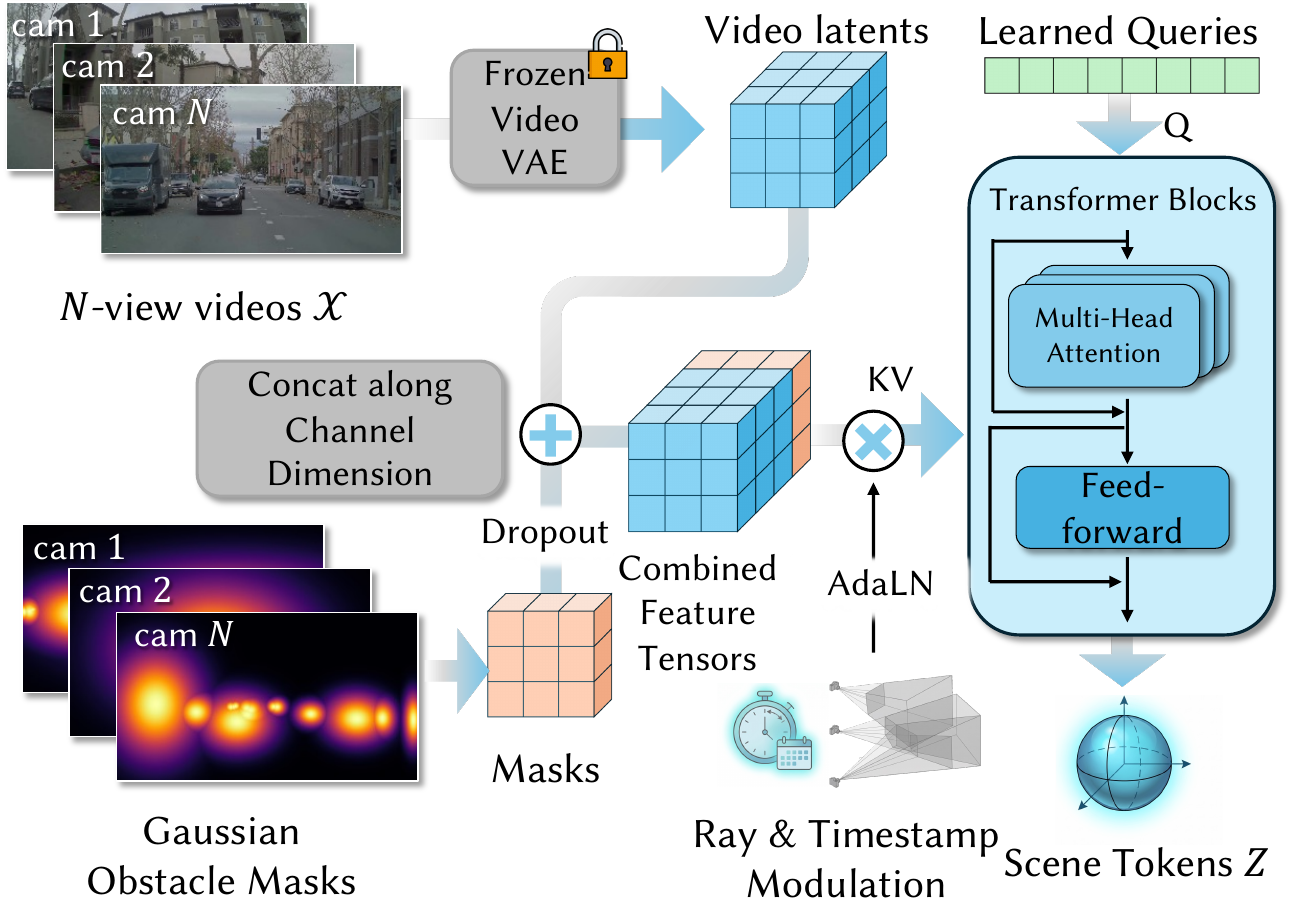}
    \caption{\textbf{Scene encoder.} A frozen video VAE extracts per-view latents from $N$ camera videos. Gaussian obstacle masks derived from projected 3D boxes are concatenated with these latents; the mask input is randomly dropped during training by setting it to zero. Camera ray and timestamp embeddings modulate the resulting features via AdaLN. Within each temporal span, a fixed number of learned queries aggregate features across cameras. Tokens from successive spans form the ordered sequence $Z$, whose length is independent of $N$ for a fixed video duration.}
    \label{fig:encoder}
    \vspace{-1.5em}
\end{figure}

\noindent\textbf{Encoder design.}
As shown in Fig.~\ref{fig:encoder}, given $N$ input camera videos $\mathcal X$, a frozen video VAE extracts features from each view. The encoder processes these features in discrete temporal spans. Within each span, camera ray and timestamp embeddings modulate the features through AdaLN~\cite{peebles2023scalable}, and a Q-Former~\cite{li2023blip} with learned queries aggregates information across cameras. The resulting tokens form a temporally ordered sequence $Z\in\mathbb R^{K\times d}$. The total token count $K$ is determined by the video duration and the query budget for each span, independently of the camera count $N$. The temporal span schedule, query budget, latent dimension $d$, and their selection rationale are provided in the supplements.

\noindent\textbf{Decoder design.}
Here, output views refer to camera streams from the camera rigs used during training. For each requested view, the decoder embeds its camera rays and timestamps to form target-view queries. Following ray-conditioned decoding in SRT~\cite{sajjadi2022scene}, we embed calibrated camera-ray maps and timestamps to form target-view queries. The ray maps specify 3D origins and directions; the queries
cross-attend to $Z$ to predict video latents, which the frozen video VAE converts into frames. To render additional views, the decoder forms new queries while reusing the same $Z$, leaving the scene-token count unchanged. We train the scene representation with reconstruction losses and a KL term on the encoder's Gaussian posterior over $Z$.

\subsection{Region-Aware Capacity Allocation}
The compact scene-token budget creates a trade-off in how much capacity can be devoted to different parts of the scene. Inspired by the importance-sampling principle~\cite{chen2025dora} used in compact 3D representations, we give greater learning emphasis to regions containing moving vehicles, whose appearance and motion are especially important for driving videos. Our implementation does not increase the token count or explicitly resample tokens; instead, Gaussian obstacle masks increase the influence of these regions on both scene encoding and reconstruction.

We project the annotated 3D vehicle boxes into each camera view. For each projected box, we form a 2D Gaussian centered at its projected center, with covariance estimated from the projected points on the box. The per-view mask is the pixelwise maximum of these Gaussians. We concatenate the masks with the corresponding video-VAE latents at the scene-encoder input (Fig.~\ref{fig:encoder}), providing an explicit cue about where the limited representation capacity should focus.

We also use the masks to reweight the reconstruction objective. During training, we randomly drop the masks only from the encoder input, while the original masks weight the RGB reconstruction term on frames decoded from the predicted video latents by the frozen video VAE:

\begin{equation}
\label{eq:obj}
\mathcal L_{\mathrm{obj}}
=\frac{\sum_{t,n,p}M_{t,n,p}\,
\ell_1(\hat x_{t,n,p},x_{t,n,p})}
{\sum_{t,n,p}M_{t,n,p}+\delta},
\end{equation}
where $n$ indexes target views, $p$ indexes pixels, $M\in[0,1]$ is the Gaussian obstacle mask, $\delta>0$ is a small constant, and
$\ell_1(\hat x,x)=\frac{1}{3}\sum_{c\in\{R,G,B\}}|\hat x_c-x_c|$
is the per-pixel RGB error. This term encourages $Z$ to retain foreground appearance. Further training details are provided in the supplementary material. In Sec.~\ref{sec:generation}, we detail the generation of these scene-token sequences.

\begin{table*}[t]
\centering
\caption{\textbf{Reconstruction quality and representation size for static and dynamic multi-camera videos.} Floats/Pixel is the total number of latent scalars divided by the total number of input pixels across cameras and frames.}
\label{tab:quantitative_results}
\begin{tabular}{l c ccc ccc}
\toprule
\multirow{2}{*}{\textbf{Method}} & \multirow{2}{*}{\textbf{Floats/Pixel $\downarrow$}} & \multicolumn{3}{c}{Static Environment} & \multicolumn{3}{c}{Dynamic Environment} \\
\cmidrule(lr){3-5} \cmidrule(lr){6-8}
 & & \textbf{PSNR} $\uparrow$ & \textbf{SSIM} $\uparrow$ & \textbf{LPIPS} $\downarrow$ & \textbf{PSNR} $\uparrow$ & \textbf{SSIM} $\uparrow$ & \textbf{LPIPS} $\downarrow$ \\
\midrule
\rowcolor{black!5} \multicolumn{8}{l}{\textit{\textcolor{black!30}{Non-queryable video representation (Ceiling)}}} \\
\rowcolor{black!5} \textcolor{black!30}{GAE~\cite{lu2026gae}} & \textcolor{black!30}{3.27e-1} & \textcolor{black!30}{34.84} & \textcolor{black!30}{0.95} & \textcolor{black!30}{0.11} & \textcolor{black!30}{34.09} & \textcolor{black!30}{0.96} & \textcolor{black!30}{0.09} \\
\rowcolor{black!5} \textcolor{black!30}{Cosmos 3 (Wan 2.2)~\cite{agarwal2026cosmos, wan2025wan}} & \textcolor{black!30}{5.12e-2} & \textcolor{black!30}{33.97} & \textcolor{black!30}{0.94} & \textcolor{black!30}{0.07} & \textcolor{black!30}{32.77} & \textcolor{black!30}{0.95} & \textcolor{black!30}{0.06} \\
\midrule
\multicolumn{8}{l}{\textit{Queryable video representation}} \\
4DGT~\cite{xu20254dgt} & 75.00 & 20.26 & 0.56 & 0.48 & 18.93 & 0.53 & 0.50 \\
MoVies~\cite{lin2026movies} & 10.74 & 25.92 & 0.77 & 0.25 & 25.15 & 0.76 & 0.24 \\
GLD~\cite{jang2026repurposing} & 7.83 & 24.57 & 0.65 & 0.33 & 23.49 & 0.64 & 0.34 \\
SceneTok~\cite{asim2026scenetok} & 1.52e-2 & 21.88 & 0.59 & 0.43 & \textendash & \textendash & \textendash \\
\textbf{Ours} & \cellcolor{myblue}\textbf{1.28e-2} & \cellcolor{myblue}\textbf{30.52} & \cellcolor{myblue}\textbf{0.86} & \cellcolor{myblue}\textbf{0.17} & \cellcolor{myblue}\textbf{29.07} & \cellcolor{myblue}\textbf{0.86} & \cellcolor{myblue}\textbf{0.16} \\
\bottomrule
\end{tabular}
\end{table*}

\begin{figure*}
    \centering
    \includegraphics[width=1.0\linewidth]{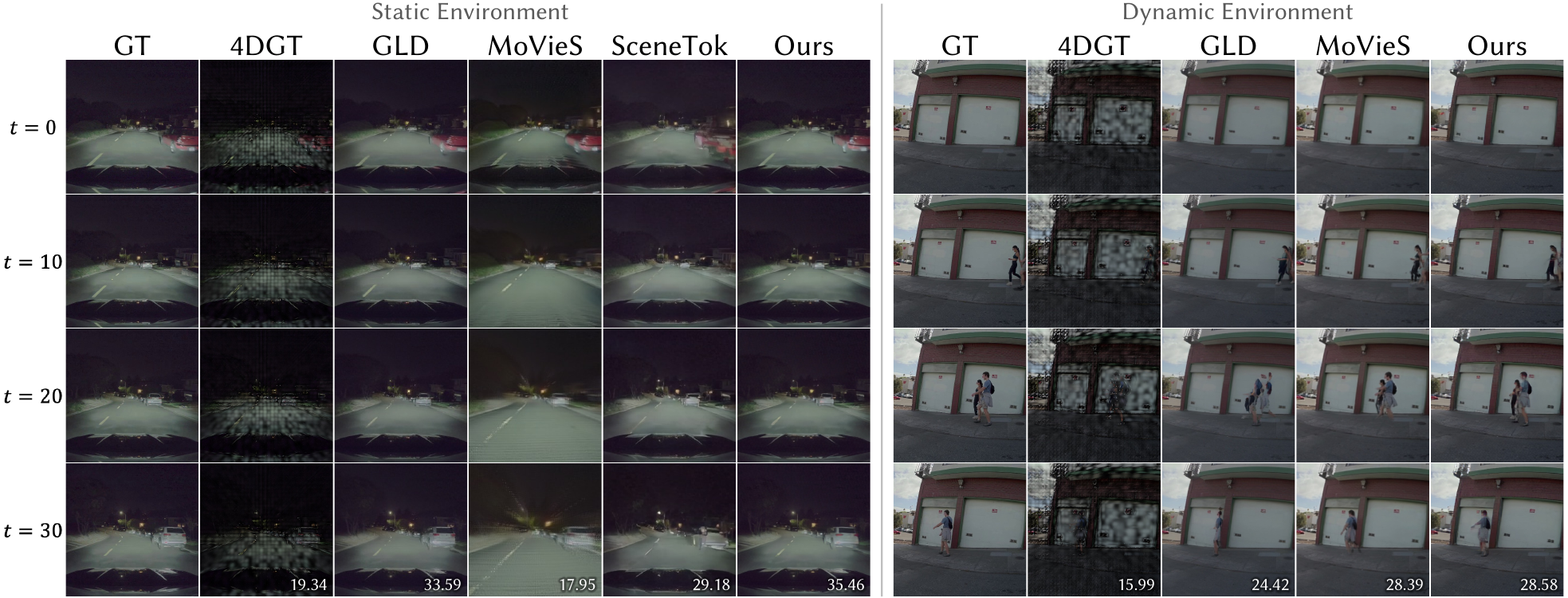}
    \caption{\textbf{Qualitative comparison of video reconstruction.} We evaluate our method against baselines on static (left) and dynamic (right) environments, visualizing four selected timesteps ($t=0,10,20,30$). The number at the bottom right of each result indicates the average PSNR over the full corresponding clip.}
    \label{fig:qualitative}
\end{figure*}

\subsection{Generating Videos with Scene Tokens}
\label{sec:generation}

With the encoder and decoder fixed, we encode each training video as a target scene-token sequence $Z$ and train a conditional flow-matching model with a DiT-style velocity predictor $v_\theta$. The condition $C$ comprises tokens encoded from the initial camera observations and embeddings of the provided 3D object boxes and class labels. Given standard Gaussian noise $\epsilon$ matching the shape of $Z$, and a continuous flow time $s\in[0,1]$, we train $v_\theta$ along the path
\begin{equation}
\label{eq:flow_matching}
\begin{aligned}
Z_s &= (1-s)Z+s\epsilon,\\
\mathcal L_{\mathrm{FM}}
&= \mathbb E_{Z,C,\epsilon,s}
   \left[\|v_\theta(Z_s,s,C)-(\epsilon-Z)\|_2^2\right].
\end{aligned}
\end{equation}

At inference, the flow model samples one scene-token sequence from noise conditioned on $C$. The decoder renders the requested views from these tokens using camera rays and timestamps. Additional views require more decoder queries, but no additional scene tokens or another flow-model sampling run. Further implementation details are provided in the supplementary material.

\section{Experiment}

We first evaluate the learned scene representation through multi-camera video reconstruction (Sec.~\ref{exp_recon}). We then assess the quality and efficiency of conditional video generation using this representation (Sec.~\ref{exp_generation}). Finally, we conduct ablation studies to examine the contributions of the main components (Sec.~\ref{ablation}).

\begin{figure*}
    \centering
    \includegraphics[width=1.0\linewidth]{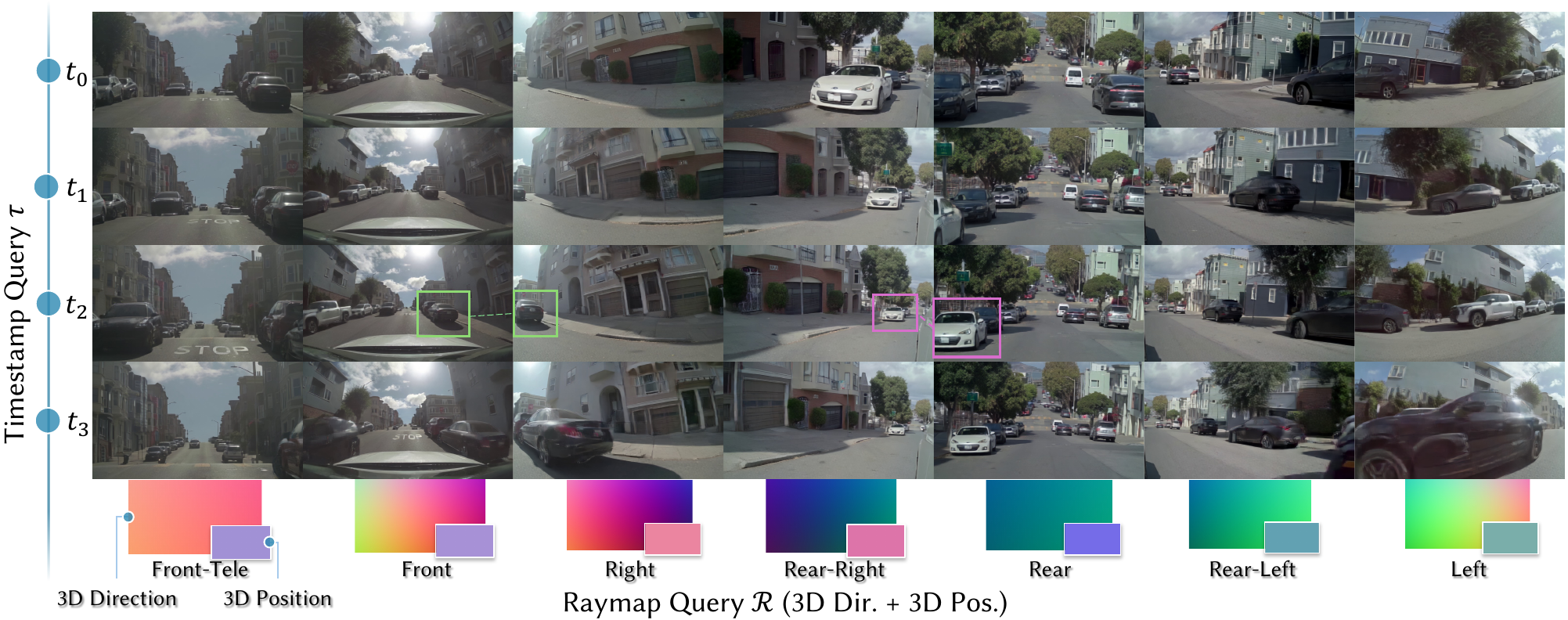}
    \caption{\textbf{Qualitative multi-view video reconstruction and camera-ray queries.} Each column is a camera view; the upper four rows show reconstructed frames at the labeled video times $t_0$–$t_3$($t=0$ -- $t=30$). At $t_2$, boxes with the same color identify the same vehicle where camera fields of view overlap. The bottom row visualizes each view's six-channel ray query: the large map shows the three ray-direction components, and the inset shows the three ray-origin (3D position) components, with each triplet displayed as RGB.}
    \label{fig:multiview}
\end{figure*}

\subsection{Experimental Setup}
\label{exp_setup}

\noindent\textbf{Baselines and Evaluation Protocol.} For reconstruction, controlled baselines include diffusable  Gaussian-based 4DGT~\cite{xu20254dgt} and MoVieS~\cite{lin2026movies}, as well as diffusable latent-based GLD~\cite{jang2026repurposing} and SceneTok~\cite{asim2026scenetok}. We report state-of-the-art per-view representations as performance ceilings, specifically GAE~\cite{lu2026gae} without temporal compression and Cosmos 3 with temporal compression. For generation, we compare diffusion sampling time and cross-view consistency with Cosmos 2.5-MV~\cite{ali2025world} and Cosmos 3-MV~\cite{agarwal2026cosmos}. These are system-level comparisons, as their model scales and pretraining data differ from ours. For overall video quality, we compare gFID and gFVD with HorizonDrive~\cite{zhang2026horizondrive} and MagicDrive-V2~\cite{gao2025magicdrive}, fine-tuning the latter on our training data and reporting model sizes alongside the results. All compared methods receive the same 3D bounding boxes for each test clip through their respective conditioning mechanisms. We train on the training split of the NVIDIA PhysicalAI Autonomous Vehicles dataset \footnote{\url{https://huggingface.co/datasets/nvidia/PhysicalAI-Autonomous-Vehicles}}, which contains 153,625 20-second training clips (approximately 853.5 hours of raw video), and evaluate on 99 clips held out from training. Unless stated otherwise, all comparisons use 33 frames from 7 cameras at 10 FPS, with a per-view resolution of $480\times864$ and matched fields of view. GenOne uses the pretrained Wan 2.2 video VAE~\cite{wan2025wan} as a frozen per-view component. Stage 1 and Stage 2 were each trained on 384 PPU-ZW810E accelerators, taking 390.3 and 195.3 hours of wall-clock time, respectively.

\noindent\textbf{Runtime and Ranking Convention.} 
All runtime measurements are conducted on NVIDIA H20 GPUs. Additional implementation details of our model are provided in the supplementary material. In our tables, blue cells mark the best controlled variants. Gray rows denote external baselines like GAE and Cosmos 3 excluded from the primary ranking, as they maintain separate per-view representations rather than others' unified queryable latent.

\begin{table}[htbp]
  \centering
    \vspace{-0.5em}
  
  \caption{\textbf{Multi-view consistency.} Quantitative comparison of query view reconstruction using the MEt3R metric.}
  \label{tab:met3r_query}
  \resizebox{\columnwidth}{!}{
  \begin{tabular}{lcc}
    \toprule
    \textbf{Method} & \textbf{Static (MEt3R $\downarrow$)} & \textbf{Dynamic (MEt3R $\downarrow$)} \\
    \midrule
    4DGT~\cite{xu20254dgt}        & 0.7558         & 0.7526         \\
    MoVies~\cite{lin2026movies}      & 0.5909         & 0.5693         \\
    GLD~\cite{jang2026repurposing}         & 0.7801             & 0.7804             \\
    SceneTok~\cite{asim2026scenetok}    & 0.6161         & --             \\
    \textbf{Ours} & \cellcolor{myblue}\textbf{0.5601} & \cellcolor{myblue}\textbf{0.5303} \\
    \bottomrule
  \end{tabular}
    \vspace{-1.5em}
  
  }
\end{table}

\subsection{Multi-view Video Reconstruction}
\label{exp_recon}

\begin{table*}[htbp]
    \centering
    \caption{\textbf{Latent token counts and GPU time for multi-camera video generation.} Sampling covers iterative diffusion or flow steps; total inference also includes conditioning and decoding. Times are summed over the GPUs used by each method. }
    \label{tab:inference_performance}
    \begin{tabular}{l c c c}
        \toprule
        \textbf{Method} & \textbf{\makecell{\# Noisy Latent \\ Tokens Num}} & \textbf{Sampling GPU Time} & \textbf{Total Inference GPU Time} \\
        \midrule
        Cosmos 2.5 (35 sampling steps) & 408k & 27\,min 2\,s & 31\,min 9\,s\\
        Cosmos 3 (35 sampling steps)   & 102k & 10\,min 48\,s & 12\,min 46\,s\\
        HorizonDrive (4 steps, w/ DMD) & 408k &  3 min 19s & 3\,min 52\,s \\ 
        \midrule
        Ours (Per-view tokens, 35 steps)$^\dagger$ & 102k & 29\,min 14\,s & 31\,min 32\,s \\ 
        Ours (35 sampling steps) & \cellcolor{myblue}\textbf{25k} &  \cellcolor{myblue}\textbf{2\,min 49\,s} & \cellcolor{myblue}\textbf{3\,min 35\,s}\\
        \bottomrule
        \multicolumn{4}{@{}l}{\footnotesize $^\dagger$ Used for runtime profiling only.}
    \end{tabular}%
\vspace{-1.5em}
\end{table*}

\noindent\textbf{Compactness and Fidelity.} Compressing multi-view videos into a unified queryable latent naturally involves a trade-off between representation compactness and spatial-temporal fidelity. Tab.~\ref{tab:quantitative_results} and Fig.~\ref{fig:qualitative} evaluate this balance quantitatively and qualitatively. Fidelity is measured across standard visual metrics (PSNR, SSIM, and LPIPS). Compactness is measured in Floats/Pixel, defined as the total number of latent floats divided by the original multi-view pixel count (i.e., $\frac{H_l \times W_l \times T_l \times C_{latent}}{H \times W \times T \times N_{\text{views}}}$).

Existing queryable baselines exhibit distinct compromises. Gaussian-based 4DGT and MoVieS maintain spatial structures but incur prohibitive storage overheads. Conversely, highly compressed methods like SceneTok lose substantially more appearance information, leading to severe quality degradation. Our method navigates this trade-off more effectively. It achieves the highest reconstruction metrics among all evaluated queryable baselines in both static and dynamic environments, while using orders of magnitude fewer Floats/Pixel than Gaussian methods and remaining even more compact than SceneTok.

\noindent\textbf{Multi-cam Consistency.} Beyond per-view fidelity, a unified latent must preserve geometric alignment across cameras. Tab.~\ref{tab:met3r_query} evaluates this per-frame multi-view consistency using MEt3R~\cite{asim2025met3r}, which directly measures symmetric feature disagreement between view pairs. Lower MEt3R scores indicate better geometric coherence.   

Quantitatively, our method achieves the lowest MEt3R errors among all baselines in both static and dynamic environments. Qualitatively, Fig.~\ref{fig:multiview} demonstrates this alignment; objects within overlapping camera regions (e.g., highlighted vehicles) remain spatially consistent across adjacent views.

The supplementary material presents longer-clip reconstructions, alongside qualitative reconstructions on held-out Waymo~\cite{sun2020waymo} and Argoverse~\cite{wilson2021argoverse} clips using Stage 1 models fine-tuned for each camera rig.

\subsection{Multi-camera Video Generation}
\label{exp_generation}

Tables~\ref{tab:inference_performance}--\ref{tab:video_quality} jointly evaluate GenOne in terms of sampling efficiency, cross-view consistency, and generation quality.

\noindent\textbf{Inference Efficiency.}
 Tab.~\ref{tab:inference_performance} reports aggregate GPU time. Experiments use a pool of eight NVIDIA H20 GPUs; for each method, the reported time is summed over the GPUs used in that run. Timing begins after model loading, with inputs available as tensors, and ends when output video tensors are produced. Dataset reading and output writing are excluded. Sampling GPU time covers the iterative diffusion or flow steps; total inference GPU time also includes the method's conditioning and decoding stages.

As shown in Tab.~\ref{tab:inference_performance}, at 35 sampling steps, GenOne has lower sampling GPU time than Cosmos 2.5-MV and Cosmos 3-MV; its sampling time is also lower than that of the four-step distilled HorizonDrive. To examine the effect of the shared representation within GenOne, we profile \textit{Ours (Per-view tokens)}, a per-view-token configuration which maintains independent view sequences. Under the same 35-step setting, this variant takes 29 min 14 s with 102k tokens, compared with 2 min 49 s with 25k shared tokens for GenOne. This within-model comparison supports the sampling benefit of the shared representation. Although the per-view variant and Cosmos 3 have similar token counts, their runtimes also reflect differences in model architecture and implementation. For a fixed clip duration, one generated scene representation serves all requested views, while rendering additional views incurs scene decoder and video VAE decoding costs.

\begin{table}[htbp]
    \centering
    \caption{\textbf{Cross-view consistency of generated videos.} MEt3R and VGGT-based depth and RGB alignment errors are lower when views agree more closely.}
    \label{tab:multiview_consistency}
    \resizebox{\columnwidth}{!}{
    \begin{tabular}{l c c c}
        \toprule
        \textbf{Method} & \textbf{MEt3R $\downarrow$} & \textbf{Depth Align. $\downarrow$} & \textbf{RGB Align. $\downarrow$} \\
        \midrule
        Cosmos 2.5~\cite{ali2025world} & 0.60 & 0.46 & 0.14 \\
        Cosmos 3~\cite{agarwal2026cosmos}   & 0.60 & 0.45 & 0.14 \\
        MagicDrive-v2~\cite{gao2025magicdrive} & 0.60 & 0.49 & 0.16 \\
        Ours       & \cellcolor{myblue}\textbf{0.58} & \cellcolor{myblue}\textbf{0.41} & \cellcolor{myblue}\textbf{0.13} \\
        \bottomrule
    \end{tabular}%
    \vspace{-1.5em}
    }
\end{table}

\begin{figure*}
    \centering
    \includegraphics[width=1.0\linewidth]{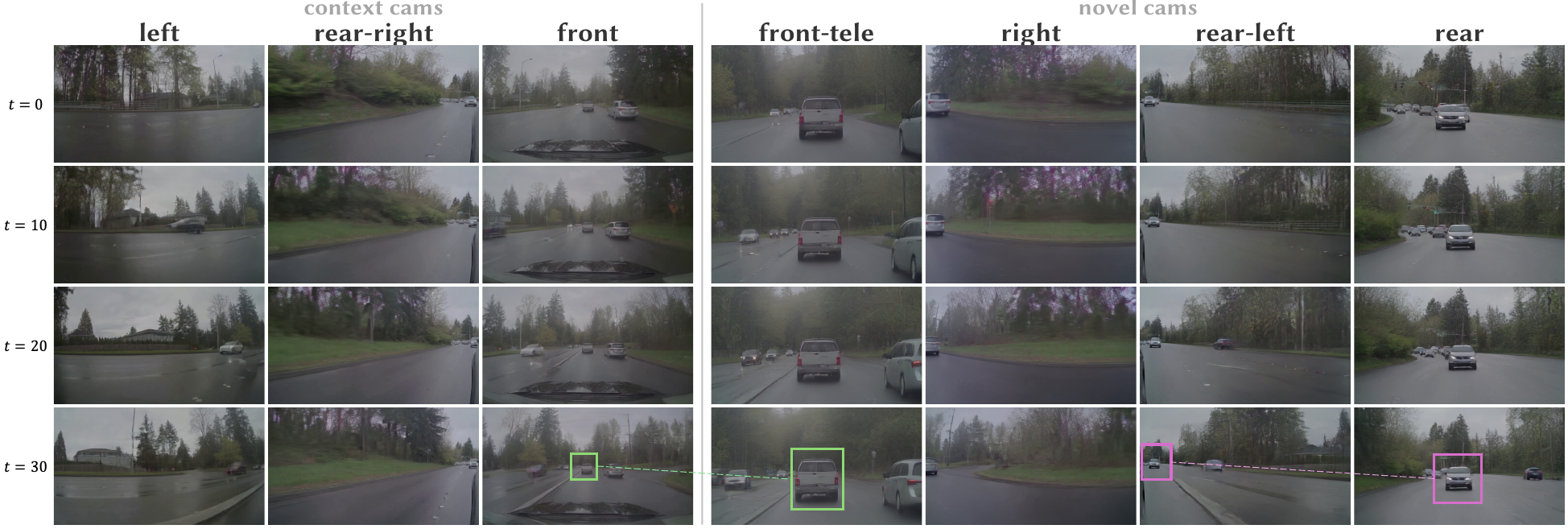}
    \caption{\textbf{Multi-camera video generation from one sampled scene representation.} The first three columns show context-camera views, and the remaining four show novel-camera views. Rows show generated frames at indices $t=0,10,20,30$. At $t=30$, \textcolor[HTML]{8ED973}{green} boxes identify the same vehicle in the front and front-tele views (context-to-novel), while \textcolor[HTML]{D86ECC}{magenta} boxes identify another vehicle in the rear-left and rear views (novel-to-novel).}
    \label{fig:generation_multiview}
\end{figure*}

\begin{table}[htbp]
\centering
\vspace{-1.5em}
\caption{\textbf{Generative video quality and model size.}}
\label{tab:video_quality}
\begin{tabular}{lccc}
\toprule
\textbf{Method} & \textbf{\# Params} & \textbf{gFID $\downarrow$} & \textbf{gFVD $\downarrow$} \\
\midrule
MagicDrive-v2~\cite{gao2025magicdrive} & 2.04B & 30.01 & 203.81 \\
HorizonDrive~\cite{zhang2026horizondrive} & 1.42B & 19.33 & 188.83 \\
Ours & 0.78B & \cellcolor{myblue}\textbf{19.25} & \cellcolor{myblue}\textbf{82.05} \\
\bottomrule
\end{tabular}
\end{table}

\noindent\textbf{Generative Video Quality.}
Tab.~\ref{tab:multiview_consistency} reports multi-view consistency using MEt3R and VGGT-based~\cite{wang2025vggt} Depth and RGB Alignment.  GenOne performs on par with Cosmos 2.5 and 3, confirming that our unified scene representation accelerates inference without compromising cross-view structural and appearance coherence. For sparse-view generation, the flow model is trained with context tokens encoded from a subset of cameras, while the target scene tokens are encoded from the full camera set. Fig.~\ref{fig:generation_multiview} shows seven views decoded from a single sampled scene latent, including four cameras absent from the three-camera context. Rendering these additional views requires neither more generated scene tokens nor another flow-model pass. 

Furthermore, to ensure a fair evaluation of overall generative quality, we benchmark GenOne (0.78B parameters) against baselines at similar scales: HorizonDrive (1.42B) and MagicDrive (2.04B, fine-tuned on our identical dataset). As shown in Tab.~\ref{tab:video_quality}, under this controlled setup, GenOne achieves comparable performance to these baselines. Results at different output resolutions are provided in the supplements.

\subsection{Ablation Studies}
\label{ablation}

\begin{figure}[t]
    \centering
    \begin{minipage}{\columnwidth}
        \centering
        \small
        \setlength{\tabcolsep}{3pt}
        \captionof{table}{\textbf{Foreground PSNR (dB) for three training-mask settings.} Higher is better.}
        \label{tab:importance_sampling_ablation}
        \begin{tabular}{lccc}
        \toprule
        \textbf{Method} & \textbf{w/o Mask} & \textbf{Binary BBox} & \textbf{Gaussian Heatmap} \\
        \midrule
        \textbf{FG PSNR$\uparrow$} & 27.96 & 29.52 & \cellcolor{myblue}\textbf{29.78} \\
        \bottomrule
        \end{tabular}
    \end{minipage}
    
    \vspace{0.3cm}
    
    \begin{minipage}{\columnwidth}
        \centering
        \includegraphics[width=\linewidth]{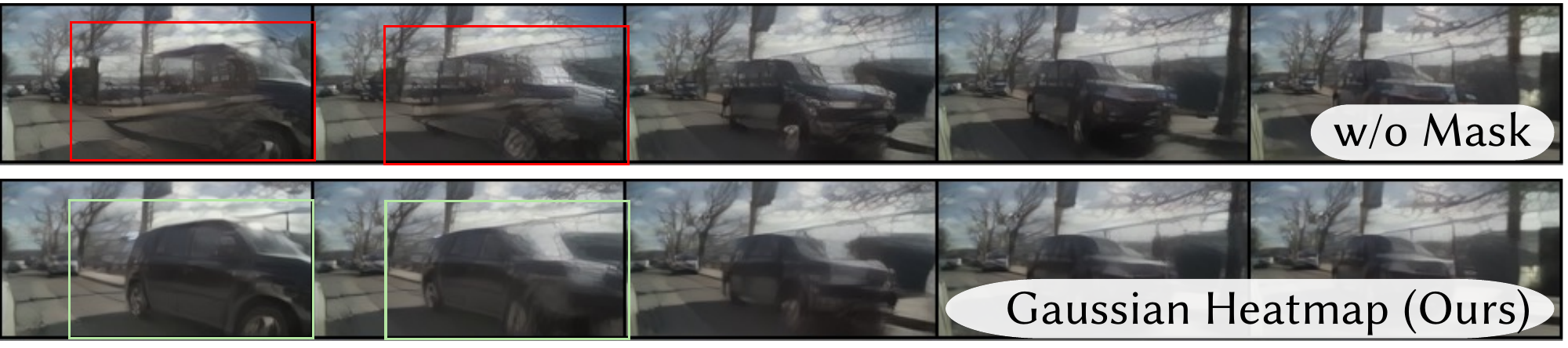}
        \caption{\textbf{Effect of Gaussian obstacle-mask supervision on a moving vehicle.} Rows compare no mask (top) and Gaussian heatmap (bottom) over successive frames. Colored boxes identify the vehicle regions shown in both reconstructions.}
        \label{fig:qualitative_foreground}
    \end{minipage}
\end{figure}

\noindent\textbf{Region-Aware Capacity Allocation.} To evaluate the effect of importance sampling and its encoding scheme on foreground regions, we conduct an ablation study with three settings: No mask, Binary BBox, and Gaussian heatmap. All variants are initialized from the same checkpoint and trained with the same fine-tuning configuration, number of updates, and validation set.

As shown in Tab.~\ref{tab:importance_sampling_ablation} and Fig.~\ref{fig:qualitative_foreground}, incorporating importance sampling improves foreground PSNR. The Binary BBox setting yields a higher PSNR than the No mask baseline, and the Gaussian heatmap achieves the highest foreground PSNR. This indicates that the Gaussian heatmap provides a representation that facilitates foreground reconstruction.

\begin{figure}
    \centering
    \includegraphics[width=1.0\linewidth]{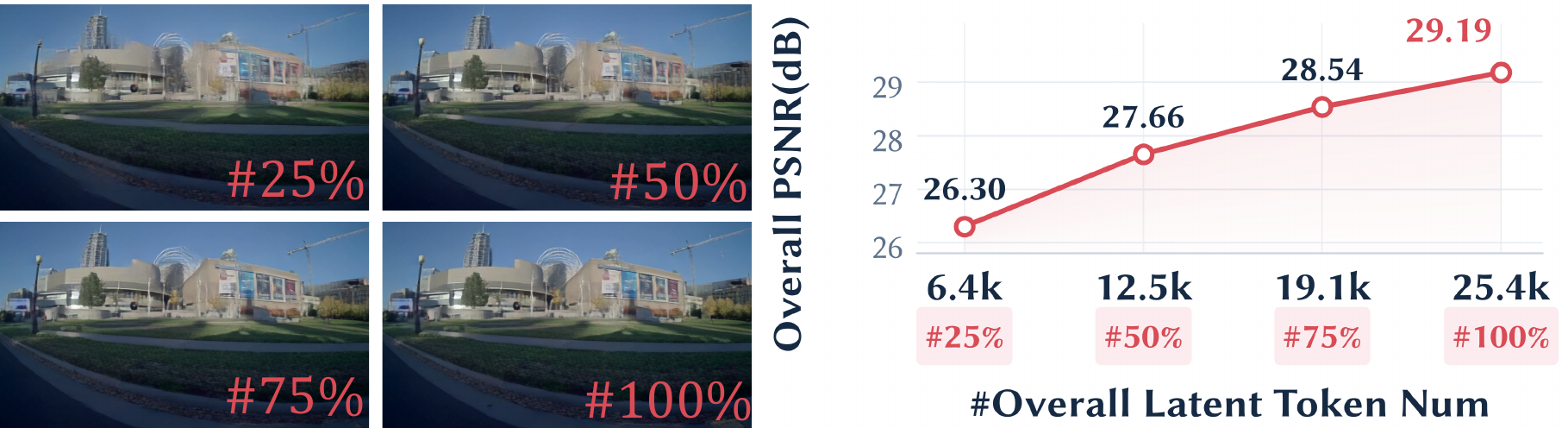}
    \caption{\textbf{Effect of the inference-time scene-token budget on reconstruction.} All settings use the same checkpoint and retain 25\%, 50\%, 75\%, or 100\% of the full query budget. Left: reconstructed frames at each budget. Right: overall PSNR versus the retained scene-token count.}
    \label{fig:tokenbudget}
\end{figure}

\noindent\textbf{Token Budget vs. Reconstruction Quality.}
Using a fixed checkpoint, we retain part of shared latent queries at inference. Relative to the full budget, PSNR decreases by $0.65$, $1.53$, and $2.89$ dB at $75\%$, $50\%$, and $25\%$, respectively. Despite these quantitative drops, qualitative results (Fig.~\ref{fig:tokenbudget}, left) show that the reconstructed scenes remain largely similar even at the $25\%$ budget.

\begin{table}[t]
  \centering
  \small
  \setlength{\tabcolsep}{5pt}
  \caption{\textbf{Inference time across output view counts.}
  The same sampled scene is rendered for every row.
  Sum includes sampling, rendering, and conditioning.}
  \begin{tabular}{@{}lrrrr@{}}
    \toprule
    & \multicolumn{4}{c}{Runtime (s)} \\
    \cmidrule(l){2-5}
    \textbf{Views} & \textbf{Sampling} & \textbf{Scene dec.}
      & \textbf{RGB dec.} & \textbf{Sum} \\
    \midrule
    1 & \textbf{168.96} & 0.71 & 6.40 & 176.08 \\
    3 & \textbf{168.96} & 0.87 & 19.21 & 189.04 \\
    5 & \textbf{168.96} & 0.99 & 32.03 & 201.98 \\
    7 & \textbf{168.96} & 1.11 & 44.84 & 214.91 \\
    \bottomrule
  \end{tabular}
  \label{tab:output_view_scaling}
\end{table}

\noindent\textbf{Output-view scaling.}
We render nested subsets of one, three, five, and seven target cameras from the same sampled scene (Tab.~\ref{tab:output_view_scaling}). The scene decoder produces per-view video latents, which the frozen video VAE converts into RGB frames. In this setting, increasing the output count from one to seven adds 38.83\,s to sampling-plus-rendering time: 38.43\,s in latent-to-RGB decoding and 0.40\,s in scene-to-latent decoding. The scene-token budget and number of flow-model passes remain fixed by construction; additional views incur rendering work.

\section{Conclusion}
We introduced GenOne, a generate-once, render-multiple-views framework for multi-camera driving video generation. GenOne learns a shared scene representation, generates it with conditional flow matching, and renders requested camera views from the same sampled representation. Experiments show competitive multi-view reconstruction, cross-view consistency, and generation quality, together with substantially faster sampling than the evaluated baselines. These results demonstrate the value of modeling a driving scene once and rendering it across views.
\clearpage
\section*{LLM Usage Statement}
The authors utilized ChatGPT and Gemini for English language editing and grammar polishing of the main text. The authors reviewed and edited all outputs as needed and take full responsibility for the ultimate content of this manuscript.
{
    \small
    \bibliographystyle{ieeenat_fullname}
    \bibliography{main}

@String(CVPR= {IEEE Conf. Comput. Vis. Pattern Recog.})

@String(ICCV= {Int. Conf. Comput. Vis.})

@String(AAAI = {AAAI})

@String(CVPR  = {CVPR})

@String(ICCV  = {ICCV})

@article{pham2026tracktok,
  title={TrackTok: Object-Centric Video Tokenization with Semantically Persistent Tokens},
  author={Pham, Nhi and Wewer, Christopher and Schiele, Bernt and Fischer, Jonas and Lenssen, Jan Eric},
  journal={arXiv preprint arXiv:2610.33857},
  year={2026}
}

@inproceedings{jin2025lvsm,
  title={Lvsm: A large view synthesis model with minimal 3d inductive bias},
  author={Jin, Haian and Jiang, Hanwen and Tan, Hao and Zhang, Kai and Bi, Sai and Zhang, Tianyuan and Luan, Fujun and Snavely, Noah and Xu, Zexiang},
  booktitle={International Conference on Learning Representations},
  volume={2025},
  pages = {60001--60021},
  year={2025}
}

@inproceedings{szymanowicz2026lagernvs,
  title={LagerNVS: Latent geometry for fully neural real-time novel view synthesis},
  author={Szymanowicz, Stanislaw and Chen, Minghao and Wang, Jianyuan and Rupprecht, Christian and Vedaldi, Andrea},
  booktitle={Proceedings of the IEEE/CVF Conference on Computer Vision and Pattern Recognition},
  pages={15443--15453},
  year={2026}
}

@inproceedings{xu20254dgt,
  title={4dgt: Learning a 4d gaussian transformer using real-world monocular videos},
  author={Xu, Zhen and Li, Zhengqin and Dong, Zhao and Zhou, Xiaowei and Newcombe, Richard and Lv, Zhaoyang},
  booktitle={Advances in neural information processing systems},
  year={2025}
}

@inproceedings{lin2026movies,
  title = {MoVieS: Motion-Aware {4D} Dynamic View Synthesis in One Second},
  author = {Lin, Chenguo and Lin, Yuchen and Pan, Panwang and Yu, Yifan and Hu, Tao and Yan, Honglei and Fragkiadaki, Katerina and Mu, Yadong},
  booktitle = {2026 IEEE/CVF Conference on Computer Vision and Pattern Recognition (CVPR)},
  year = {2026},
  pages = {295--306},
}

@inproceedings{jang2026repurposing,
  title={Repurposing geometric foundation models for multi-view diffusion},
  author={Jang, Wooseok and Jeon, Seonghu and Han, Jisang and Choi, Jinhyeok and Kwon, Minkyung and Kim, Seungryong and Xie, Saining and Liu, Sainan},
  booktitle={European Conference on Computer Vision},
  pages={609--627},
  year={2026},
  organization={Springer}
}

@inproceedings{asim2026scenetok,
  title     = {SceneTok: A Compressed, Diffusable Token Space for 3D Scenes},
  author    = {Asim, Mohammad and Wewer, Christopher and Lenssen, Jan Eric},
  booktitle = {2026 IEEE/CVF Conference on Computer Vision and Pattern Recognition (CVPR)},
  pages= {5870-5880},
  year      = {2026}
}

@inproceedings{gao2025magicdrive,
  title={Magicdrive-v2: High-resolution long video generation for autonomous driving with adaptive control},
  author={Gao, Ruiyuan and Chen, Kai and Xiao, Bo and Hong, Lanqing and Li, Zhenguo and Xu, Qiang},
  booktitle={2025 IEEE/CVF International Conference on Computer Vision (ICCV)},
  pages={28135--28144},
  year={2025},
  organization={IEEE}
}

@article{agarwal2026cosmos,
  title={Cosmos 3: Omnimodal world models for physical ai},
  author={Agarwal, Niket and Ali, Arslan and Allen, Jon and Antolini, Martin and Aubame, Adeline and Azzolini, Alisson and Bai, Junjie and Bala, Maciej and Balaji, Yogesh and Bapst, Josh and others},
  journal={arXiv preprint arXiv:2606.02800},
  year={2026}
}

@article{zhang2026horizondrive,
  title={HorizonDrive: Self-Corrective Autoregressive World Model for Long-horizon Driving Simulation},
  author={Zhang, Conglang and Zhan, Yifan and Wang, Qingjie and Ouyang, Zhanpeng and Li, Yu and Yang, Zihao and Guo, Xiaoyang and Ren, Weiqiang and Zhang, Qian and Dong, Zhen and others},
  journal={arXiv preprint arXiv:2605.11596},
  year={2026}
}

@article{zhang2024bevworld,
  title={Bevworld: A multimodal world model for autonomous driving via unified bev latent space},
  author={Zhang, Yumeng and Gong, Shi and Xiong, Kaixin and Ye, Xiaoqing and Tan, Xiao and Wang, Fan and Huang, Jizhou and Wu, Hua and Wang, Haifeng},
  journal={arXiv preprint arXiv:2407.05679},
  year={2024}
}

@article{xu2026driveunitok,
  title={DriveUniTok: A Unified Multimodal Tokenizer for Driving World-Action Models},
  author={Xu, Jingbo and Leng, Hanchao and Huang, JiaHui and Li, Hanbing and Zhao, Jingwei and Yang, Kuiyuan and Gao, Chen and Chen, Guang and Ye, Hangjun and Li, Yong},
  journal={arXiv preprint arXiv:2610.35044},
  year={2026}
}

@article{wan2025wan,
  title={Wan: Open and advanced large-scale video generative models},
  author={Wan, Team and Wang, Ang and Ai, Baole and Wen, Bin and Mao, Chaojie and Xie, Chen-Wei and Chen, Di and Yu, Feiwu and Zhao, Haiming and Yang, Jianxiao and others},
  journal={arXiv preprint arXiv:2503.20314},
  year={2025}
}

@article{ali2025world,
  title={World simulation with video foundation models for physical ai},
  author={Ali, Arslan and Bai, Junjie and Bala, Maciej and Balaji, Yogesh and Blakeman, Aaron and Cai, Tiffany and Cao, Jiaxin and Cao, Tianshi and Cha, Elizabeth and Chao, Yu-Wei and others},
  journal={arXiv preprint arXiv:2511.00062},
  year={2025}
}

@article{yu2024repa,
  title={Representation alignment for generation: Training diffusion transformers is easier than you think},
  author={Yu, Sihyun and Kwak, Sangkyung and Jang, Huiwon and Jeong, Jongheon and Huang, Jonathan and Shin, Jinwoo and Xie, Saining},
  journal={arXiv preprint arXiv:2410.06940},
  year={2024}
}

@inproceedings{peebles2023scalable,
  title={Scalable diffusion models with transformers},
  author={Peebles, William and Xie, Saining},
  booktitle={2023 IEEE/CVF International Conference on Computer Vision (ICCV)},
  pages={4172--4182},
  year={2023},
  organization={IEEE}
}

@article{lu2026gae,
  title={GAE: Learning a Geometry-Native Latent Space for 3D-Consistent World Generation},
  author={Lu, Jiahao and Yin, Minghao and Hu, Wenbo and Liu, Hengyu and Zhao, Wang and Yeung, Sai-Kit and Shan, Ying and Liu, Yuan},
  journal={arXiv preprint arXiv:2609.24981},
  year={2026}
}

@article{chen2025bev,
  title={BEV-VAE: Multi-view Image Generation with Spatial Consistency for Autonomous Driving},
  author={Chen, Zeming and Zhao, Hang},
  journal={arXiv preprint arXiv:2507.00707},
  year={2025}
}

@inproceedings{asim2025met3r,
  title={Met3r: Measuring multi-view consistency in generated images},
  author={Asim, Mohammad and Wewer, Christopher and Wimmer, Thomas and Schiele, Bernt and Lenssen, Jan Eric},
  booktitle={2025 IEEE/CVF Conference on Computer Vision and Pattern Recognition (CVPR)},
  pages={6034--6044},
  year={2025},
  organization={IEEE}
}

@inproceedings{wang2025vggt,
  title={Vggt: Visual geometry grounded transformer},
  author={Wang, Jianyuan and Chen, Minghao and Karaev, Nikita and Vedaldi, Andrea and Rupprecht, Christian and Novotny, David},
  booktitle={2025 IEEE/CVF Conference on Computer Vision and Pattern Recognition (CVPR)},
  pages={5294--5306},
  year={2025},
  organization={IEEE}
}

@article{basant2026omnidreams,
  title={NVIDIA OmniDreams: Real-Time Generative World Model for Closed-Loop Autonomous Vehicle Simulation},
  author={Basant, Aarti and Kar, Amlan and Paschalidou, Despoina and Wei, Fangyin and Ferroni, Francesco and Cobo, Guillermo Garcia and Turki, Haithem and Ling, Huan and Seo, Jaewoo and Lucas, James and others},
  journal={arXiv preprint arXiv:2606.03159},
  year={2026}
}

@article{hwang2024emma,
  title={Emma: End-to-end multimodal model for autonomous driving},
  author={Hwang, Jyh-Jing and Xu, Runsheng and Lin, Hubert and Hung, Wei-Chih and Ji, Jingwei and Choi, Kristy and Huang, Di and He, Tong and Covington, Paul and Sapp, Benjamin and others},
  journal={Transactions on Machine Learning Research},
  year={2025}
}

@inproceedings{wang2025drivegen,
  title={Drive\&gen: Co-evaluating end-to-end driving and video generation models},
  author={Wang, Jiahao and Yang, Zhenpei and Bai, Yijing and Li, Yingwei and Zou, Yuliang and Sun, Bo and Kundu, Abhijit and Lezama, Jose and Huang, Luna Yue and Zhu, Zehao and others},
  booktitle={2025 IEEE/RSJ International Conference on Intelligent Robots and Systems (IROS)},
  pages={8934--8941},
  year={2025},
  organization={IEEE}
}

@inproceedings{yang2023unisim,
  title={Unisim: A neural closed-loop sensor simulator},
  author={Yang, Ze and Chen, Yun and Wang, Jingkang and Manivasagam, Sivabalan and Ma, Wei-Chiu and Yang, Anqi Joyce and Urtasun, Raquel},
  booktitle={2023 IEEE/CVF Conference on Computer Vision and Pattern Recognition (CVPR)},
  pages={1389--1399},
  year={2023},
  organization={IEEE}
}

@inproceedings{li2024drivingdiffusion,
  title={Drivingdiffusion: Layout-guided multi-view driving scenarios video generation with latent diffusion model},
  author={Li, Xiaofan and Zhang, Yifu and Ye, Xiaoqing},
  booktitle={European Conference on Computer Vision},
  pages={469--485},
  year={2024},
  organization={Springer}
}

@inproceedings{wen2024panaceaplus,
  title={Panacea: Panoramic and controllable video generation for autonomous driving},
  author={Wen, Yuqing and Zhao, Yucheng and Liu, Yingfei and Jia, Fan and Wang, Yanhui and Luo, Chong and Zhang, Chi and Wang, Tiancai and Sun, Xiaoyan and Zhang, Xiangyu},
  booktitle={Proceedings of the IEEE/CVF Conference on Computer Vision and Pattern Recognition},
  pages={6902--6912},
  year={2024}
}

@inproceedings{zhao2025drivedreamer,
  title={Drivedreamer-2: Llm-enhanced world models for diverse driving video generation},
  author={Zhao, Guosheng and Wang, Xiaofeng and Zhu, Zheng and Chen, Xinze and Huang, Guan and Bao, Xiaoyi and Wang, Xingang},
  booktitle={Proceedings of the AAAI Conference on Artificial Intelligence},
  volume={39},
  number={10},
  pages={10412--10420},
  year={2025}
}

@article{zheng2026xworld16,
  title={X-World: Controllable Ego-Centric Multi-Camera World Models for Scalable End-to-End Driving},
  author={Zheng, Chaoda and Li, Sean and Deng, Jinhao and Wang, Zhennan and Chen, Shijia and Xiao, Liqiang and Chi, Ziheng and Lin, Hongbin and Chen, Kangjie and Wang, Boyang and others},
  journal={arXiv preprint arXiv:2603.19979},
  year={2026}
}

@article{zhu2026xiaomiworld2,
  title={Xiaomi Auto World Model 2.0: Causal World Models for Real-World Driving and Embodied Intelligence},
  author={Zhu, Zhenxin and Huang, Zhijian and Chi, Cheng and Wang, Yuru and Wu, Jianhua and Guan, Jiayi and Li, Yonggang and Tu, Mingfei and Hu, JiaKui and Xu, Tao and others},
  journal={arXiv preprint arXiv:2611.01677},
  year={2026}
}

@article{cheng2026m4world,
  title={M4 World: A Multi-view Multimodal Driving World Model for Interactive Object Manipulation and Minute-long Streaming},
  author={Cheng, Ke and Ye, Hanqiao and Shi, Lei and Liu, Yahui and Shen, Yunhan and Dong, Jingtao and Wang, Zhenke and Ao, Wenxuan and Xu, Weixiang and Huang, Kaining and others},
  journal={arXiv preprint arXiv:2607.14005},
  year={2026}
}

@article{russell2025gaia,
  title={Gaia-2: A controllable multi-view generative world model for autonomous driving},
  author={Russell, Lloyd and Hu, Anthony and Bertoni, Lorenzo and Fedoseev, George and Shotton, Jamie and Arani, Elahe and Corrado, Gianluca},
  journal={arXiv preprint arXiv:2503.20523},
  year={2025}
}

@inproceedings{wu2025drivescape,
  title={Drivescape: High-resolution driving video generation by multi-view feature fusion},
  author={Wu, Wei and Guo, Xi and Tang, Weixuan and Huang, Tingxuan and Wang, Chiyu and Ding, Chenjing},
  booktitle={2025 IEEE/CVF Conference on Computer Vision and Pattern Recognition (CVPR)},
  pages={17187--17196},
  year={2025},
  organization={IEEE}
}

@article{ren2025cosmosdrivedreams,
  title={Cosmos-drive-dreams: Scalable synthetic driving data generation with world foundation models},
  author={Ren, Xuanchi and Lu, Yifan and Cao, Tianshi and Gao, Ruiyuan and Huang, Shengyu and Sabour, Amirmojtaba and Shen, Tianchang and Pfaff, Tobias and Wu, Jay Zhangjie and Chen, Runjian and others},
  journal={arXiv preprint arXiv:2506.09042},
  year={2025}
}

@article{zhang2026anyscene,
  title={AnyScene: Towards Highly Controllable Driving Scene Generation at Anywhere and Beyond},
  author={Zhang, Haiming and Zhou, Junfei and Jiang, Feng and Li, Jingzhong and Guo, Zhenglong and Dai, Penglin and Dai, Jifeng and Xie, Yan and Zhu, Benjin},
  journal={arXiv preprint arXiv:2605.26113},
  year={2026}
}

@inproceedings{wang2024drivewm,
  title={Driving into the future: Multiview visual forecasting and planning with world model for autonomous driving},
  author={Wang, Yuqi and He, Jiawei and Fan, Lue and Li, Hongxin and Chen, Yuntao and Zhang, Zhaoxiang},
  booktitle={2024 IEEE/CVF Conference on Computer Vision and Pattern Recognition (CVPR)},
  pages={14749--14759},
  year={2024},
  organization={IEEE}
}

@inproceedings{lu2024wovogen,
  title={Wovogen: World volume-aware diffusion for controllable multi-camera driving scene generation},
  author={Lu, Jiachen and Huang, Ze and Yang, Zeyu and Zhang, Jiahui and Zhang, Li},
  booktitle={European conference on computer vision},
  pages={329--345},
  year={2024},
  organization={Springer}
}

@inproceedings{li2023blip,
  title={Blip-2: Bootstrapping language-image pre-training with frozen image encoders and large language models},
  author={Li, Junnan and Li, Dongxu and Savarese, Silvio and Hoi, Steven},
  booktitle={International conference on machine learning},
  pages={19730--19742},
  year={2023},
  organization={PmLR}
}

@inproceedings{sun2020waymo,
  title={Scalability in perception for autonomous driving: Waymo open dataset},
  author={Sun, Pei and Kretzschmar, Henrik and Dotiwalla, Xerxes and Chouard, Aurelien and Patnaik, Vijaysai and Tsui, Paul and Guo, James and Zhou, Yin and Chai, Yuning and Caine, Benjamin and others},
  booktitle={2020 IEEE/CVF Conference on Computer Vision and Pattern Recognition (CVPR)},
  pages={2443--2451},
  year={2020},
  organization={IEEE}
}

@inproceedings{wilson2021argoverse,
  title={Argoverse 2: Next Generation Datasets for Self-Driving Perception and Forecasting},
  author={Wilson, Benjamin and Qi, William and Agarwal, Tanmay and Lambert, John and Singh, Jagjeet and Khandelwal, Siddhesh and Pan, Bowen and Kumar, Ratnesh and Hartnett, Andrew and Pontes, Jhony Kaesemodel and Ramanan, Deva and Carr, Peter and Hays, James},
  booktitle={Proceedings of the Neural Information Processing Systems Track on Datasets and Benchmarks},
  year={2021}
}

@inproceedings{sajjadi2022scene,
  title={Scene representation transformer: Geometry-free novel view synthesis through set-latent scene representations},
  author={Sajjadi, Mehdi SM and Meyer, Henning and Pot, Etienne and Bergmann, Urs and Greff, Klaus and Radwan, Noha and Vora, Suhani and Lu{\v{c}}i{\'c}, Mario and Duckworth, Daniel and Dosovitskiy, Alexey and others},
  booktitle={2022 IEEE/CVF Conference on Computer Vision and Pattern Recognition (CVPR)},
  pages={6219--6228},
  year={2022},
  organization={IEEE}
}

@article{zhuo2026drivetok,
  title={DriveTok: 3D Driving Scene Tokenization for Unified Multi-View Reconstruction and Understanding},
  author={Zhuo, Dong and Zheng, Wenzhao and Zuo, Sicheng and Yan, Siming and Hou, Lu and Zhou, Jie and Lu, Jiwen},
  journal={arXiv preprint arXiv:2603.19219},
  year={2026}
}

@inproceedings{jiang2025rayzer,
  title={Rayzer: A self-supervised large view synthesis model},
  author={Jiang, Hanwen and Tan, Hao and Wang, Peng and Jin, Haian and Zhao, Yue and Bi, Sai and Zhang, Kai and Luan, Fujun and Sunkavalli, Kalyan and Huang, Qixing and others},
  booktitle={2025 IEEE/CVF International Conference on Computer Vision (ICCV)},
  pages={4918--4929},
  year={2025},
  organization={IEEE}
}

@article{jiang2026dive,
  title={Dive: Efficient multi-view driving scenes generation based on video diffusion transformer},
  author={Jiang, Junpeng and Hong, Gangyi and Zhang, Miao and Hu, Hengtong and Zhan, Kun and Shao, Rui and Nie, Liqiang},
  journal={IEEE Transactions on Multimedia},
  year={2026},
  publisher={IEEE}
}

@article{li2026far,
  title={FAR-Drive: Frame-AutoRegressive Video Generation in Closed-Loop Autonomous Driving},
  author={Li, Yaoru and Landi, Federico and Godi, Marco and Jin, Xin and Fu, Ruiju and Ma, Yufei and Sun, Muyang and Si, Heyu and Guo, Qi},
  journal={arXiv preprint arXiv:2603.14938},
  year={2026}
}

@inproceedings{wei20263drae,
  title     = {Any 3D Scene is Worth 1K Tokens: 3D-Grounded Representation for Scene Generation at Scale},
  author    = {Wei, Dongxu and Xu, Qi and Li, Zhiqi and Zhou, Hangning and Qiu, Cong and Qin, Hailong and Yang, Mu and Cui, Zhaopeng and Liu, Peidong},
  booktitle = {Advances in Neural Information Processing Systems},
  year      = {2026}
}

@inproceedings{zhao2025wavelet,
  title={Wavelet-gs: 3d gaussian splatting with wavelet decomposition},
  author={Zhao, Beizhen and Zhou, Yifan and Yu, Sicheng and Wang, Zijian and Wang, Hao},
  booktitle={Proceedings of the 33rd ACM International Conference on Multimedia},
  pages={8616--8625},
  year={2025}
}

@inproceedings{chen2025dora,
  title={Dora: Sampling and benchmarking for 3d shape variational auto-encoders},
  author={Chen, Rui and Zhang, Jianfeng and Liang, Yixun and Luo, Guan and Li, Weiyu and Liu, Jiarui and Li, Xiu and Long, Xiaoxiao and Feng, Jiashi and Tan, Ping},
  booktitle={2025 IEEE/CVF Conference on Computer Vision and Pattern Recognition (CVPR)},
  pages={16251--16261},
  year={2025},
  organization={IEEE}
}

@inproceedings{wang2024dust3r,
  title={Dust3r: Geometric 3d vision made easy},
  author={Wang, Shuzhe and Leroy, Vincent and Cabon, Yohann and Chidlovskii, Boris and Revaud, Jerome},
  booktitle={2024 IEEE/CVF Conference on Computer Vision and Pattern Recognition (CVPR)},
  pages={20697--20709},
  year={2024},
  organization={IEEE}
}

@article{oquab2023dinov2,
  title={DINOv2: Learning Robust Visual Features without Supervision},
  author={Oquab, Maxime and Darcet, Timoth{\'e}e and others},
  journal={Transactions on Machine Learning Research},
  year={2024}
}

@inproceedings{szegedy2016pool3,
  title={Rethinking the inception architecture for computer vision},
  author={Szegedy, Christian and Vanhoucke, Vincent and Ioffe, Sergey and Shlens, Jon and Wojna, Zbigniew},
  booktitle={Proceedings of the IEEE conference on computer vision and pattern recognition},
  pages={2818--2826},
  year={2016}
}

@inproceedings{carreira2017i3d,
  title={Quo vadis, action recognition? a new model and the kinetics dataset},
  author={Carreira, Joao and Zisserman, Andrew},
  booktitle={2017 IEEE conference on computer vision and pattern recognition (CVPR)},
  pages={4724--4733},
  year={2017},
  organization={IEEE}
}
}

\clearpage
\setcounter{page}{1}
\maketitlesupplementary

This supplementary material is organized as follows. Sec.~\ref{sec:implementation} describes the scene representation and training, multi-camera video generation, and evaluation and runtime details. Sec.~\ref{sec:recon_results} covers longer-clip reconstruction and reconstruction on Waymo and Argoverse. Sec.~\ref{sec:gen_results} covers generation at different output resolutions and more qualitative image to video genration results. Sec.~\ref{sec:limitations} clarifies the scope of the sampling-cost comparison, limitations and future works.

\section{Implementation Details}
\label{sec:implementation}

\subsection{Scene Representation Configuration}
The frozen Wan~2.2 video VAE~\cite{wan2025wan} maps each camera stream to 48-channel latents, downsampling space by 16 and time by 4. Let $T_{\mathrm{lat}}$ be the number of latent time steps. The scene encoder groups the first step on its own and the remaining steps in consecutive pairs. For the clip lengths used here, $T_{\mathrm{lat}}$ is odd, giving $G=1+(T_{\mathrm{lat}}-1)/2$ groups. In group $g$, $K_g$ learned queries aggregate features from all $N$ input cameras. Concatenating the group outputs in temporal order yields $Z\in\mathbb{R}^{K\times d}$, where $K=\sum_{g=1}^{G}K_g$. At a fixed clip length, $K$ is independent of $N$.

\begin{table}[ht]
\centering
\small
\begin{tabular*}{\columnwidth}{@{\extracolsep{\fill}}ll@{}}
\hline
\textbf{Component} & \textbf{Configuration} \\
\hline
Video VAE & 48 ch.; spatial $/16$, temporal $/4$ \\
Scene encoder & 12 blocks; width 768; 12 heads \\
Scene decoder & 12 blocks; width 768; 12 heads \\
Queries per group & 2,835 (first); 5,670 (later) \\
Scene tokens & $25{,}515\times48$ (33 frames) \\
\hline
\end{tabular*}
\caption{Scene representation settings at $480\times864$.}
\label{tab:scene_config}
\end{table}

We set the shared-token budget to about one quarter of seven separate video-VAE streams at the same clip length and resolution, guiding the choice of $K$ and $d$. At this budget, the shared tokens contain fewer scalars than separate video-VAE latents for two or more views. We allocate queries in proportion to the latent steps per group: each two-step group has twice the first group's budget. We select $d=48$ because the KL term remained stable and non-negligible during Stage~1 training. For a 33-frame clip, the video VAE produces nine latent steps, grouped as $(1)$, $(2,3)$, $(4,5)$, $(6,7)$, and $(8,9)$. With $K_1=2{,}835$, $K_g=5{,}670$ for $g=2,\ldots,5$, and $d=48$, the representation has $K=25{,}515$ tokens. With $H,W,T,N$ denoting frame height, width, length, and camera count, it uses $Kd/(HWTN)=1.28\times10^{-2}$ floats per input pixel for seven cameras. Seven separate video-VAE latent streams use $5.12\times10^{-2}$ floats per pixel under the same accounting. Four frame-aligned obstacle-mask maps are packed as additional channels at each video-VAE latent step before scene encoding. The decoder forms target queries by embedding the calibrated ray maps and adding their timestamp embeddings.

\subsection{Scene Representation Training}
The scene encoder predicts a diagonal Gaussian posterior over $Z$. We sample from it during training and optimize the scene encoder and decoder while keeping the video VAE frozen. The Stage~1 generator objective is
\begin{equation}
\begin{aligned}
\mathcal{L}_{\mathrm{Stage1}}={}&3\mathcal{L}_{\mathrm{rec}}+2\mathcal{L}_{\mathrm{LPIPS}}+3\mathcal{L}_{\mathrm{obj}}+2\mathcal{L}_{\mathrm{Haar}}\\
&+0.15\mathcal{L}_{\mathrm{GAN}}+1.0\times10^{-7}\mathcal{L}_{\mathrm{KL}}.
\end{aligned}
\end{equation}
Here $\mathcal{L}_{\mathrm{rec}}$ is RGB squared error over valid pixels, $\mathcal{L}_{\mathrm{LPIPS}}$ is VGG-based perceptual error, and $\mathcal{L}_{\mathrm{obj}}$ is the foreground-weighted RGB error in Eq.~(\ref{eq:obj}). $\mathcal{L}_{\mathrm{Haar}}$ measures $L_1$ error in the high-frequency bands of a one-level Haar decomposition~\cite{zhao2025wavelet}, while $\mathcal{L}_{\mathrm{GAN}}$ is the PatchGAN generator loss and the discriminator uses a masked hinge loss. $\mathcal{L}_{\mathrm{KL}}$ regularizes the scene-token posterior toward a standard normal prior. The KL and adversarial coefficients are annealed to the values shown. We selected the Stage~1 loss weights using layer-wise analysis with the Multiple Gradient Descent Algorithm (MGDA).

The encoder receives Gaussian obstacle maps as input cues; the foreground losses use the projected box regions. During training, the encoder mask input is set to zero for 10\% of samples while the foreground losses retain their masks. Reconstruction inference uses zero mask input. Table~6 evaluates the complete mask configurations, changing the encoder input and foreground supervision together and reporting foreground PSNR.

To train the decoder across active token budgets, 50\% of samples retain a per-slot query prefix drawn log-uniformly from 128 tokens to the full slot width, rounded to a multiple of eight; the other samples use all queries. Figure~8 applies different prefixes to one trained checkpoint to measure reconstruction sensitivity to the retained token count.

Stage~1 uses 8-bit AdamW with a final learning rate of $3\times10^{-5}$; training progresses from $96\times192$ to $480\times864$ and from single frames to clips of up to 129 frames.

\subsection{Multi-camera Video Generation}
In Stage~2, we freeze the trained scene encoder and decoder and train the conditional flow model. Its DiT has 20 blocks with hidden width 1,536, 12 attention heads, and 48-dimensional scene-token inputs. Table~5's 0.78B parameter count includes the DiT and the layers encoding 3D object boxes and class labels; the frozen scene encoder, scene decoder, and video VAE are excluded. We train with flow matching and a representation-alignment term,
\begin{equation}
\mathcal{L}_{\mathrm{gen}}=\mathcal{L}_{\mathrm{FM}}+0.05\mathcal{L}_{\mathrm{REPA}},
\end{equation}
where $\mathcal{L}_{\mathrm{FM}}$ is defined in Eq.~(\ref{eq:flow_matching}). For REPA~\cite{yu2024repa}, features after the 10th DiT block are projected from 1,536 to 48 dimensions and aligned with stop-gradient scene-token targets. Training uses AdamW at $8\times10^{-4}$ and samples clip lengths $T\in\{9,17,25,33\}$ with probabilities $\{0.2,0.2,0.2,0.4\}$. Both training stages use cosine learning-rate decay and BF16 mixed precision.

For 33-frame generation, the visual context is the frame at $t=0$ from three cameras: left, front, and rear-right (Fig.~\ref{fig:generation_multiview}). The frozen scene encoder maps these images to the first group of condition tokens. These are concatenated with the noisy target scene tokens, so both sets interact through DiT self-attention; the velocity prediction is read out from the target tokens only. The object condition provides 3D boxes and class labels throughout the clip, including future timestamps. We project the boxes into the calibrated cameras and encode their class, projected geometry, and rays within each box. Observations of an object are pooled across cameras into a token at each latent time step. Gated cross-attention lets each target scene token attend to object tokens at the same latent time step. The decoder uses rays and timestamps derived from the provided target-camera poses. The reported task is therefore synthesis from initial images under a specified future object layout and target-camera trajectory.

The two conditions are dropped independently with probability 0.1 during training. At inference, compositional classifier-free guidance combines unconditional, object-only, and full-condition velocity predictions:
\begin{equation}
v_{\mathrm{guide}}=v_U+w_o(v_O-v_U)+w_s(v_{SO}-v_O).
\end{equation}
Here $v_U$, $v_O$, and $v_{SO}$ are velocity predictions with neither condition, with object controls only, and with both conditions, respectively. For the reported setting, $w_o=3.0$ and $w_s=1.5$. Sampling uses 35 DPM-Solver++(2M) steps. The frozen scene decoder then renders the sampled tokens at the requested rays and timestamps.

\subsection{Evaluation and Runtime Details}
For multi-view video reconstruction, Table~\ref{tab:quantitative_results} reports PSNR, SSIM, and LPIPS alongside floats per pixel, while Table~\ref{tab:met3r_query} reports MEt3R separately for static and dynamic scenes. For multi-camera video generation, the evaluation in Tables~3--5 and the output-view runtime in Table~\ref{tab:output_view_scaling} use the settings described below.

We evaluate all clips in the generation evaluation set. Each clip has 33 frames at 10 FPS, seven cameras, and $480\times864$ resolution. The target views, including the four absent from the visual context, belong to the camera rig used in training. Labels such as $t=10,20,30$ in the result figures denote frame indices, corresponding to 1, 2, and 3 seconds after the initial frame.

Table~\ref{tab:inference_performance} reports aggregate GPU time, summed over the GPUs used by each method and averaged over clips. Sampling time covers the iterative diffusion or flow steps; total inference time also includes conditioning and decoding. Cosmos~2.5-MV and Cosmos~3-MV use eight NVIDIA H20 GPUs each; HorizonDrive, GenOne, and the per-view-token profile use one H20 GPU each. The per-view profile processes $7\times9\times30\times54=102{,}060$ noisy tokens at 35 steps, versus 25,515 shared tokens for GenOne.

Table~\ref{tab:multiview_consistency} uses the shared output window available from all compared methods. Each generated view is center-cropped to a square and bilinearly resized to $224\times224$. At each evaluated frame, the frozen VGGT model processes all seven synchronized views jointly. We score six overlapping pairs: left--front, right--front, front--front-tele, rear-left--rear, rear-right--rear, and rear-left--rear-right. VGGT~\cite{wang2025vggt} scores both directions of each pair. The MEt3R~\cite{asim2025met3r} column uses the same pairs with pairwise DUSt3R~\cite{wang2024dust3r} geometry, DINOv2~\cite{oquab2023dinov2} cosine features.

For the VGGT columns, predicted depths and cameras project source pixels into the paired view. A correspondence is valid when both depths are positive, the projected point lies inside the target image and in front of its camera, and the projected values are finite; a direction contributes a score when at least 32 pixels are valid. Depth alignment is the median of $2|d_{\mathrm{proj}}-d_{\mathrm{target}}|/(d_{\mathrm{proj}}+d_{\mathrm{target}})$ over valid pixels, while RGB alignment is the mean absolute RGB difference after reprojection. We average common finite scores over frames and directed pairs within each clip, then average the clip scores over the scenes shared by the compared methods.

For Table~\ref{tab:video_quality}, gFID and gFVD compare generated and reference feature distributions over all evaluation clips and all seven camera views, with one generated sample per clip. Both exclude the conditioning frame at $t=0$ and use the complete remaining 32-frame video window. gFID pools frame features across cameras and time using the clean-fid Inception2015 pool3~\cite{szegedy2016pool3} extractor. gFVD pools video features across camera streams using a Kinetics-400-pretrained I3D extractor~\cite{carreira2017i3d}. Each score is a single Fr\'echet distance over the pooled generated and reference features, rather than an average of per-camera scores.

Table~\ref{tab:output_view_scaling} reports the same aggregate-GPU-time convention on one H20 GPU. Each row renders a different number of views from the same sampled scene; only scene decoding and video-VAE decoding increase with view count. Timings begin with tensor inputs after model loading and exclude dataset reading and output-file writing.

GenOne uses a frozen pretrained Wan~2.2 video VAE~\cite{wan2025wan} and trains its scene representation and conditional flow model on the NVIDIA PhysicalAI training split (Secs.~\ref{sec:implementation}.1--\ref{sec:implementation}.3). Among the multi-camera baselines, MagicDrive-V2 is fine-tuned on the same NVIDIA PhysicalAI training split and evaluated on both cross-view consistency (Table~\ref{tab:multiview_consistency}) and generative video quality (Table~\ref{tab:video_quality}). Cosmos~2.5-MV and Cosmos~3-MV are evaluated as pretrained multi-view systems with different model scales and pretraining data. HorizonDrive is evaluated in Table~\ref{tab:multiview_consistency} (4-step DMD sampling) and Table~\ref{tab:video_quality} as a single-view reference baseline rather than a multi-camera generator. At evaluation, all compared methods receive the same 3D bounding boxes for each test clip through their respective conditioning interfaces.

\section{Additional Reconstruction Results}
\label{sec:recon_results}
\subsection{Longer-Clip Reconstruction}

\begin{figure*}
    \centering
    \includegraphics[width=1.0\linewidth]{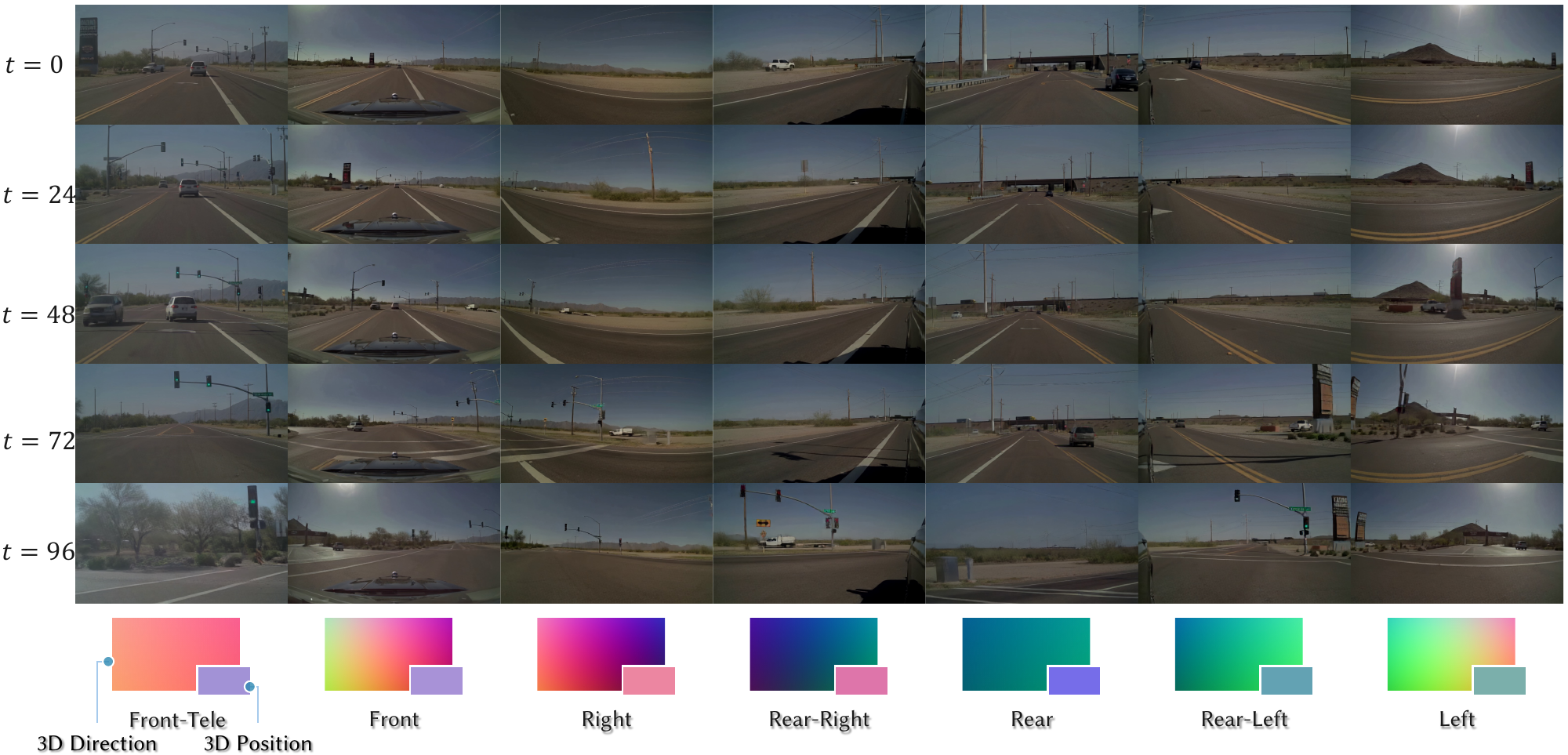}
    \caption{\textbf{Longer-clip 7-camera reconstruction (daytime).} Reconstructed frames at $t \in \{0, 24, 48, 72, 96\}$ across all seven cameras, with corresponding 3D ray direction and position maps (bottom).}
    \label{fig:longer_clip1}
\end{figure*}

\begin{figure*}
    \centering
    \includegraphics[width=1.0\linewidth]{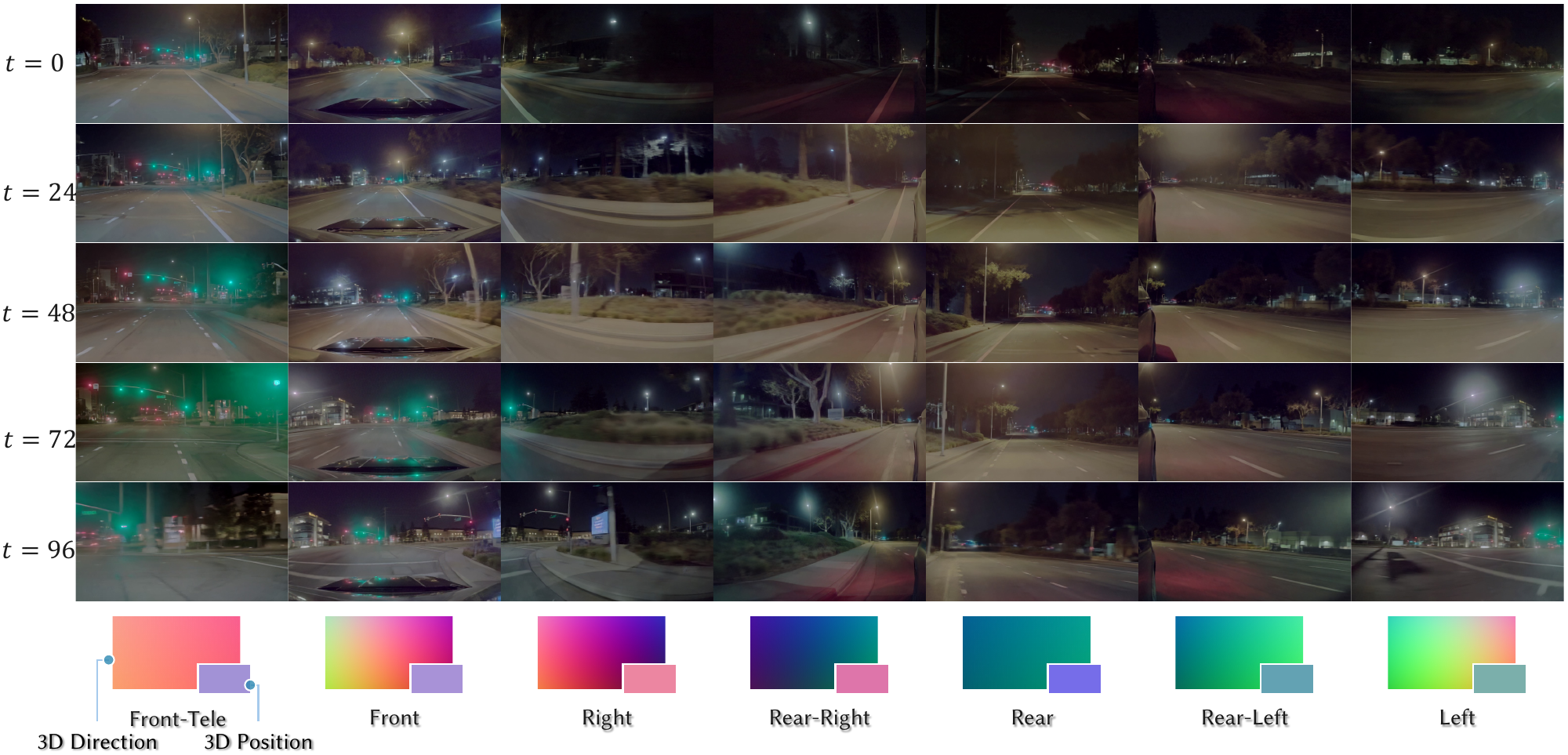}
    \caption{\textbf{Longer-clip 7-camera reconstruction (nighttime).} Reconstructed frames at $t \in \{0, 24, 48, 72, 96\}$ across all seven cameras.}
    \label{fig:longer_clip2}
\end{figure*}

Figures~\ref{fig:longer_clip1} and~\ref{fig:longer_clip2} show qualitative 7-camera reconstruction examples up to $t = 96$ frames across daytime and nighttime driving scenes. These visualizations illustrate how GenOne's shared scene representation and ray-conditioned rendering extend to longer sequences.

\begin{figure*}
    \includegraphics[width=1.0\linewidth]{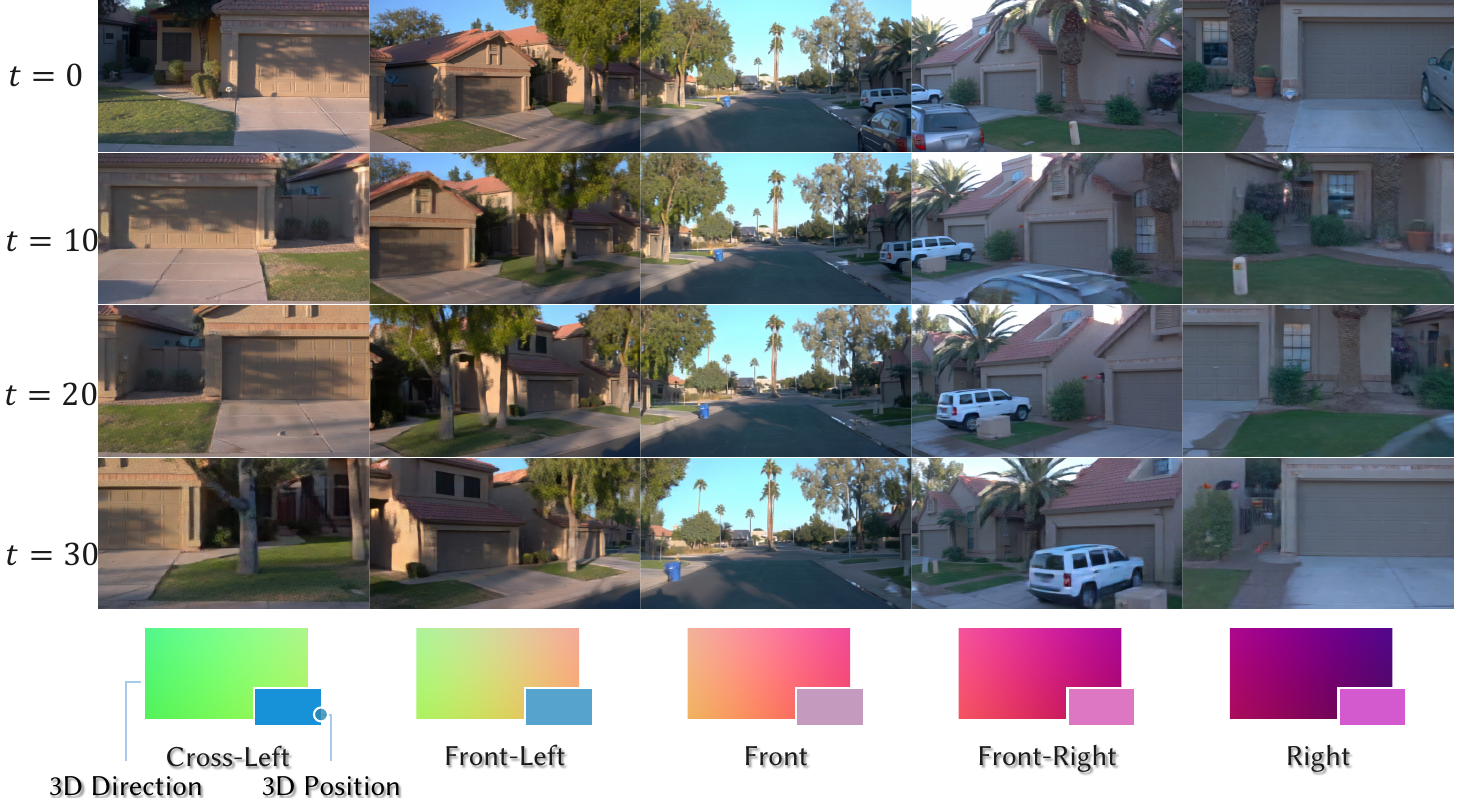}
    \caption{\textbf{Multi-camera reconstruction on Waymo.}}
    \label{fig:waymo_recon}
\end{figure*}

\begin{figure*}
    \includegraphics[width=1.0\linewidth]{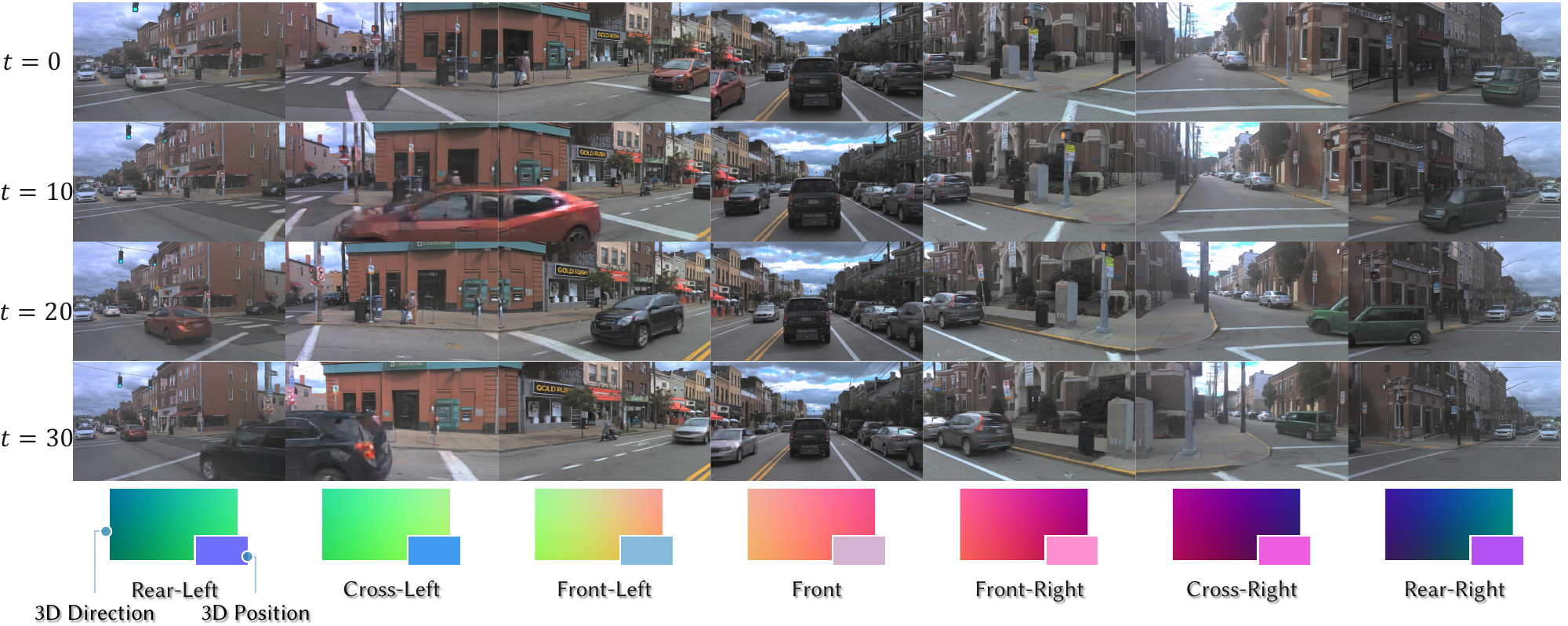}
    \caption{\textbf{Multi-camera reconstruction on Argoverse.}}
    \label{fig:argo_recon}
\end{figure*}

\subsection{Reconstruction on Waymo and Argoverse}
\label{sec:waymo_argo}
To illustrate cross-dataset adaptability, we fine-tune the Cosmos-initialized checkpoint on the Waymo Open Dataset~\cite{sun2020waymo} and Argoverse~\cite{wilson2021argoverse} for 2,500 steps. Figures~\ref{fig:waymo_recon} and~\ref{fig:argo_recon} show qualitative multi-camera reconstruction examples on both datasets.

\section{Generation Results}
\label{sec:gen_results}
\subsection{Generation Results on Different Resolution}
\begin{figure*}
    \includegraphics[width=1.0\linewidth]{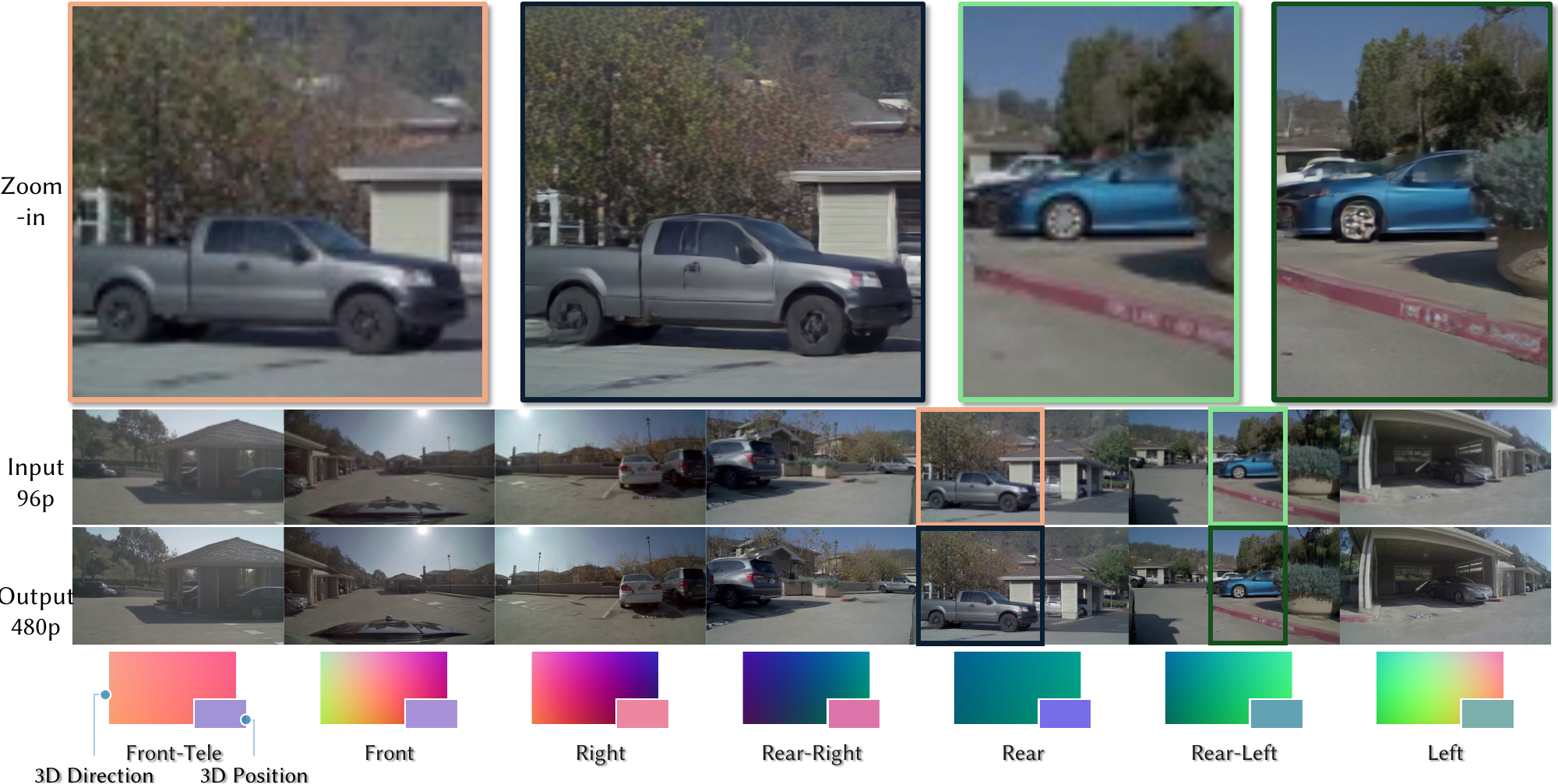}
   \caption{\textbf{96p-to-480p multi-camera decoding.} Seven-camera 480p outputs decoded from 96p inputs, with zoomed-in comparisons (top) and corresponding 3D ray direction and position maps (bottom).}
    \label{fig:resolution}
\end{figure*}

Given 96p multi-camera inputs, we encode each clip into shared scene tokens once. To decode at 480p, we rescale the camera intrinsics and construct target-view ray maps on the 480p video-VAE latent grid while keeping camera poses and timestamps fixed. The same scene tokens are then decoded using the frozen scene decoder and video VAE. Qualitative results are shown in Figure~\ref{fig:resolution}.

\subsection{Generation Results}
\begin{figure*}
    \includegraphics[width=1.0\linewidth]{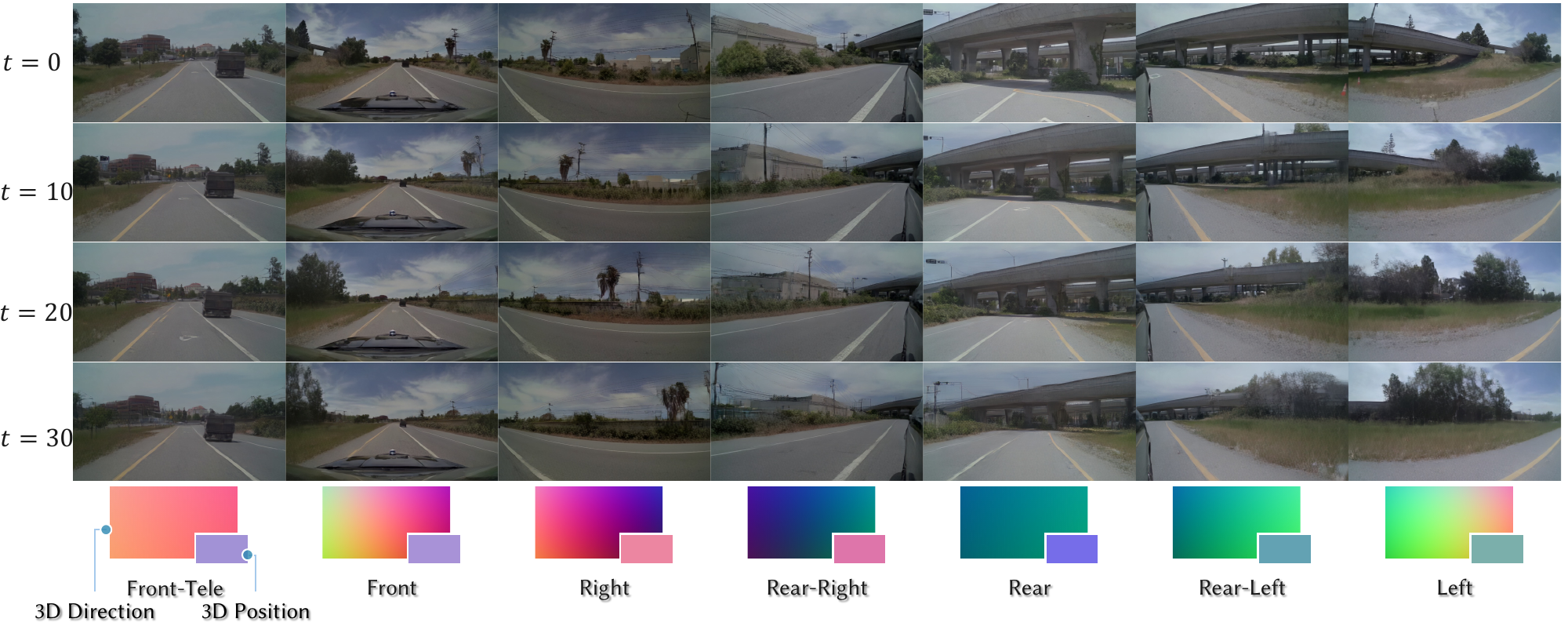}
   \caption{\textbf{7-camera video generation (daytime).}}
    \label{fig:gen_clip1}
\end{figure*}

\begin{figure*}
    \includegraphics[width=1.0\linewidth]{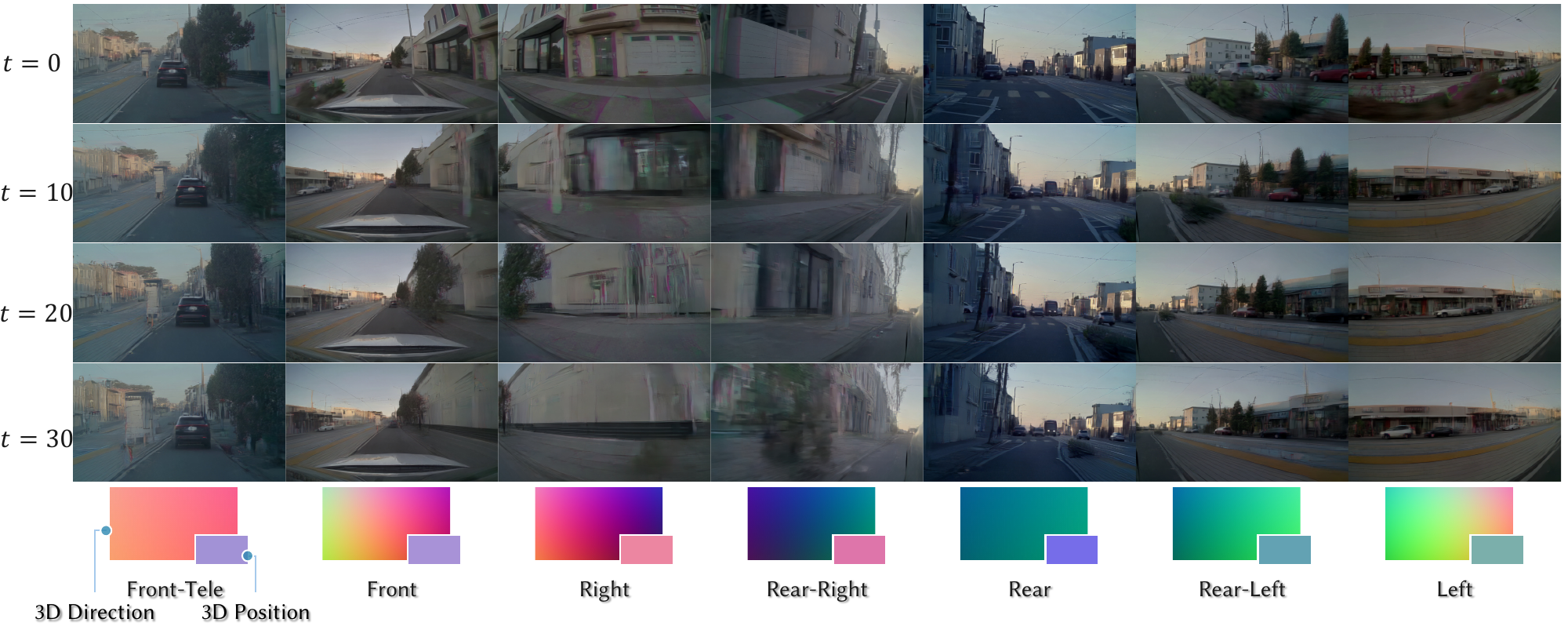}
   \caption{\textbf{7-camera video generation (golden hour).}}
    \label{fig:gen_clip2}
\end{figure*}

\begin{figure*}
    \includegraphics[width=1.0\linewidth]{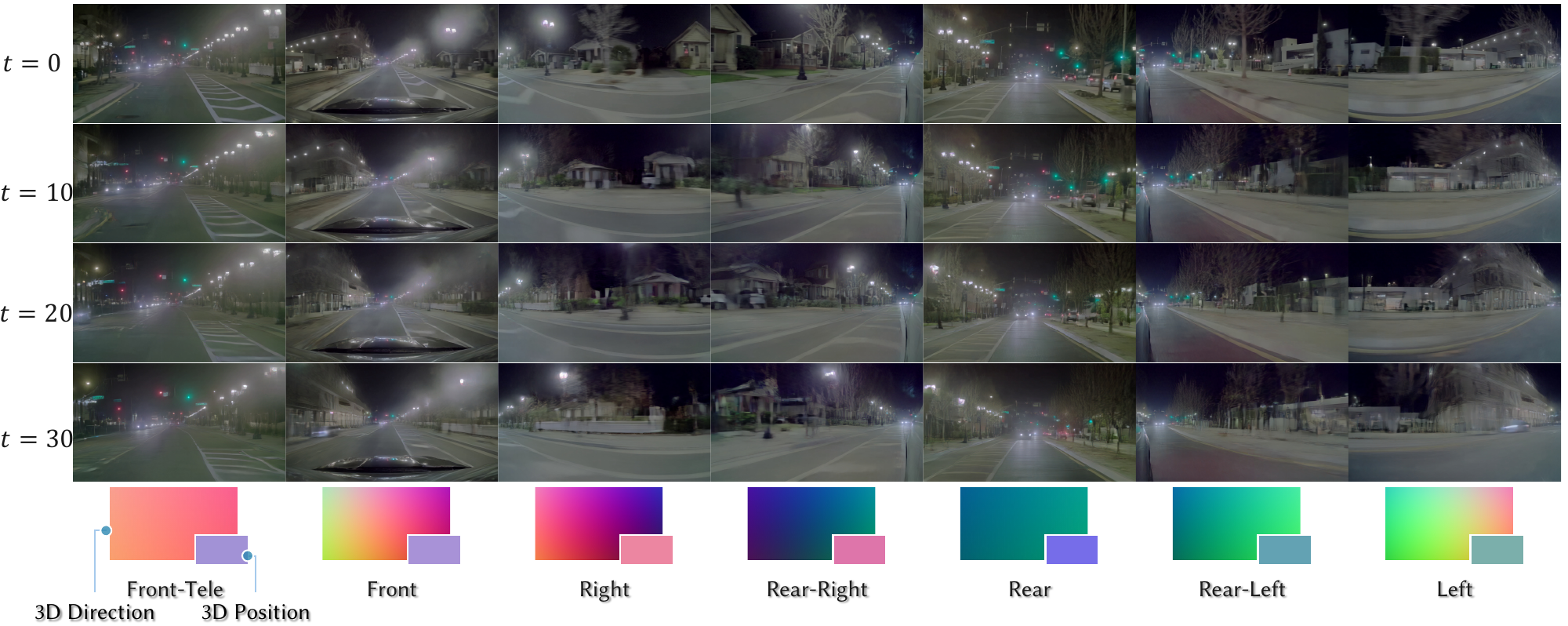}
   \caption{\textbf{7-camera video generation (nighttime).}}
    \label{fig:gen_clip3}
\end{figure*}

\begin{figure*}
    \includegraphics[width=1.0\linewidth]{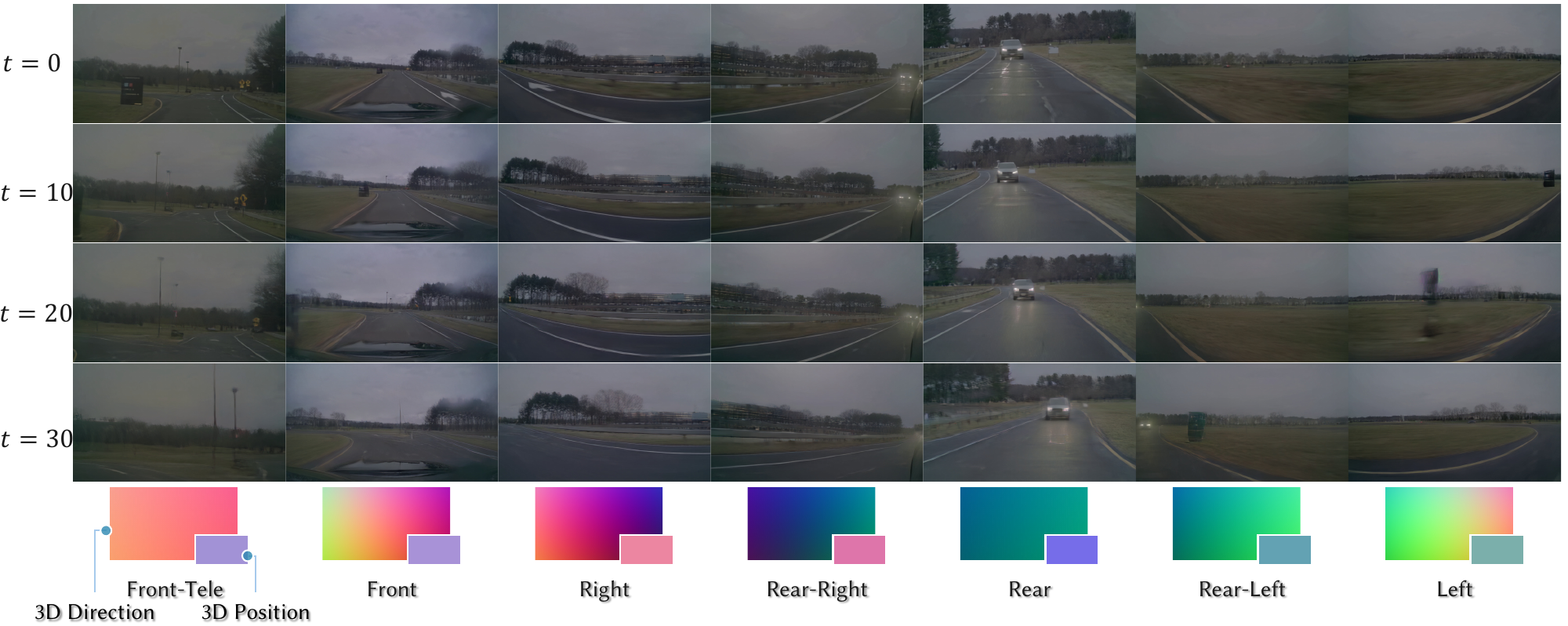}
   \caption{\textbf{7-camera video generation (overcast).} Additional video demos are available at \url{https://youtu.be/bYqQFN6etOo}.}
    \label{fig:gen_clip4}
\end{figure*}

Figures~\ref{fig:gen_clip1}--\ref{fig:gen_clip4} show additional qualitative 7-camera video generation examples across diverse driving scenes and lighting conditions.

\section{Limitations}
\label{sec:limitations}
\subsection{Scope of the Sampling-Cost and Quality Comparisons}
Table~\ref{tab:inference_performance} measures 35-step sampling cost on one NVIDIA H20 GPU at the respective per-view and shared-token counts. The profiled per-view configuration uses the same 0.78B-parameter DiT backbone as GenOne, with a sequence length of 102,060 tokens versus 25,515 shared tokens (10.4$\times$ faster sampling). Because Stage~2 training of the shared-token model required 384 PPU-ZW810E accelerators for 195.3 hours (74{,}995.2 GPU-hours), the per-view configuration is profiled for sampling cost only and is not trained to convergence. Consequently, our experiments establish the within-model computational saving of the shorter shared-token sequence, but do not isolate the generation-quality change attributable to replacing per-view tokens with shared tokens at a matched compute budget. Separately, Tables~\ref{tab:multiview_consistency} and~\ref{tab:video_quality} evaluate the trained GenOne system against external generators at the system level. Although Sec.~4.3 refers to the Table~5 benchmark as a controlled setup in terms of shared test clips, generator scale reporting, and fine-tuning MagicDrive-V2 on our training split, the compared systems still differ in architecture, pretraining history, and conditioning mechanisms. In particular, MagicDrive-V2 serves as our primary trained multi-camera baseline evaluated across both cross-view consistency (Table~\ref{tab:multiview_consistency}) and video quality (Table~\ref{tab:video_quality}), while Cosmos~2.5-MV and Cosmos~3-MV provide system-level multi-view consistency references. Taken together, the existing experiments support two distinct conclusions: (1) shared scene tokens substantially reduce within-model sampling cost, and (2) the resulting trained GenOne system achieves competitive multi-view consistency and video quality relative to established multi-camera generators under our evaluation protocol.

\subsection{Limitations and Future Work}
Our current experiments focus on real-world driving datasets with fixed camera rigs, calibrated camera parameters, and 3D bounding-box annotations. Incorporating simulation data with diverse and flexible camera placements is a natural direction for exploring broader camera-rig generalization. In addition, our Stage~2 flow model is evaluated on 33-frame clips conditioned on specified future object layouts and target-camera trajectories; extending GenOne to longer-horizon, streaming, and closed-loop rollouts remains an important direction for future work.

\end{document}